\documentclass{article}

\usepackage{relsize}
\usepackage{microtype}
\usepackage{graphicx}
\usepackage{subfigure}
\usepackage{booktabs} % for professional tables
\usepackage{xspace}
\usepackage{multirow}
\usepackage{adjustbox}
\newcommand{\mc}[3]{\multicolumn{#1}{#2}{#3}}
\newcommand{\ocd}{OC20}
\newcommand{\model}{eSCN}
\usepackage{hyperref}

\usepackage[accepted]{icml2023}

\usepackage{amsmath}
\usepackage{amssymb}
\usepackage{mathtools}
\usepackage{amsthm}

\usepackage[capitalize,noabbrev]{cleveref}

\theoremstyle{plain}
\newtheorem{theorem}{Theorem}[section]
\newtheorem{proposition}[theorem]{Proposition}

\theoremstyle{definition}
\newtheorem{definition}[theorem]{Definition}

\theoremstyle{remark}

\usepackage[textsize=tiny]{todonotes}

\icmltitlerunning{Reducing SO(3) Convolutions to SO(2) for Efficient Equivariant GNNs}

\begin{document}

\twocolumn[
\icmltitle{
Reducing SO(3) Convolutions to SO(2) for Efficient Equivariant GNNs 
}



\icmlsetsymbol{equal}{*}

\begin{icmlauthorlist}
\icmlauthor{Saro Passaro}{meta}
\icmlauthor{C. Lawrence Zitnick}{meta}

\end{icmlauthorlist}

\icmlcorrespondingauthor{Saro Passaro}{saro00@meta.com}
\icmlcorrespondingauthor{Larry Zitnick}{zitnick@meta.com}

\icmlaffiliation{meta}{Fundamental AI Research at Meta AI}




\icmlkeywords{Machine Learning, ICML}

\vskip 0.3in
]



\printAffiliationsAndNotice{} % otherwise use the standard text.

\begin{abstract}

Graph neural networks that model 3D data, such as point clouds or atoms, are typically desired to be $SO(3)$ equivariant, i.e., equivariant to 3D rotations. Unfortunately equivariant convolutions, which are a fundamental operation for equivariant networks, increase significantly in computational complexity as higher-order tensors are used.
In this paper, we address this issue by reducing the $SO(3)$ convolutions or tensor products to mathematically equivalent convolutions in $SO(2)$ . This is accomplished by aligning the node embeddings' primary axis with the edge vectors, which sparsifies the tensor product and reduces the computational complexity from $O(L^6)$ to $O(L^3)$, where $L$ is the degree of the representation. 
We demonstrate the potential implications of this improvement by proposing the Equivariant Spherical Channel Network (eSCN), a graph neural network utilizing our novel approach to equivariant convolutions, which achieves state-of-the-art results on the large-scale OC-20 and OC-22 datasets.

\end{abstract}

\section{Introduction}

In many domains, it is desired that machine learning models obey specific constraints imposed by the task. A common constraint is equivariance to specific operations on the inputs \citep{bronstein2021geometric}. That is, if the input is transformed in a certain manner, such as translated or rotated, the output should be transformed appropriately. Prominent examples include equivariance to translations for object detection in images \cite{girshick2014rich}, and equivariance to rotations of 3D point clouds \cite{weiler20183d, satorras2021n}. Machine learning models impose equivariance to a group of symmetries by constraining the operations that can be performed. Namely, the class of functions that can be learned is narrowed to those that are equivariant. The hope is that equivariance will provide a robust prior that can increase data efficiency, improve generalization and eliminate the need for data augmentation \citep{reisert2009spherical}. 

$SO(3)$-Equivariant graph neural networks (GNNs) have showed great promise in processing geometrical information, such as 3D point clouds of objects \cite{weiler20183d, chen2021equivariant} or atoms \cite{batzner20223, thomas2018tensor, brandstetter2022geometric, liao2023equiformer, musaelian2023learning}. These networks incorporate the inherent symmetry in their domains to $SO(3)$, the group of 3D rotations. Specifically, they take advantage of geometric tensors of irreducible representations, so-called irreps, to act as node embeddings and utilize directional information.  The building block of these networks is an approach to message passing that uses equivariant convolutions based on tensor products of the irreps. Unfortunately, the full tensor product using irreps up to degree $L$ have a computational complexity of $O(L^6)$, which significantly limits their use with degrees higher than 2 or 3. 

In this paper, we address this issue by proposing an efficient method to perform equivariant convolutions. Our main observation is that node irreps exhibit special properties if the irreps' primary axis is aligned to the edge's direction during message passing. While SCN \citep{zitnick2022spherical} observed that this leads to a subset of the coefficients becoming $SO(3)$ invariant, we demonstrate the benefits extend even further. Specifically, the tensor product becomes sparse, which reduces its computational complexity from $O(L^6)$ to $O(L^3)$, and removes the need to compute the Clebsch-Gordan coefficients. This enables computationally efficient equivariant models that use irreps of significantly higher degree.

Additionally, we shed light on this novel approach by revealing its relationship with $SO(2)$ convolutions \citep{worrall2017harmonic}. In fact, while predicting per-atom forces in atomic systems is a $SO(3)$-equivariant task, during message passing the symmetry of the problem gets reduced to $SO(2)$. Specifically, by aligning the irreps' primary axis with the edge's direction, only a single degree of rotational freedom remains. Thus, the $SO(3)$ irreps can be projected to the analogous irreducible representations of $SO(2)$ and the much more computationally efficient $SO(2)$ convolutions may be performed. In conclusion, the tensor product may be viewed as a set of generalized convolutions in $SO(2)$ if the irreps are rotated appropriately.

We use these insights to propose the Equivariant Spherical Channel Network (\model), an equivariant GNN utilizing the efficient implementation of the equivariant convolutions. Empirically, we evaluate our model to the task of predicting atomic energies and forces, a foundational problem in chemistry and material science with numerous important applications; including tackling climate change~\cite{zitnick2020introduction, rolnick2022tackling}. Typically, atomic forces and energies are estimated using Density Functional Theory \citep{hohenberg1964inhomogeneous, kohn1965self}, a very computationally expensive calculation; thus, the goal is to approximate DFT through Machine learning models to significantly reduce this cost. 
We compare~\model~to state-of-the-art GNNs on the large-scale OC-20 and OC-22 datasets \citep{OC20}, which contain over 100 million atomic structure training examples for catalysts to help address climate change. \model~achieves state-of-the-art performance across many tasks, especially those such as force prediction ($9\%$ and $21\%$ improvement for OC-20 and OC-22) and relaxed structure prediction ($15\%$ improvement) that require high directional fidelity.

\section{Related work}

Incorporating symmetries of the data in machine learning models can improve their data-efficiency and ability to generalize \citep{cohen2016group}. 
Group theory is a formalism to axiomatically define symmetries \citep{bronstein2021geometric}. A group $G$ is a set of elements with an operation $\bullet_G : G \times G \rightarrow G$. It acts on a vector space $V$ (i.e. $\mathbb{R}^n$ for some $n$) with an operation $* : G \times V \rightarrow V$ called group action.
We say that $F: V \rightarrow W$ is:
\begin{enumerate}
    \item \textbf{$G$-Invariant} if $\forall \mathbf{v} \in V $,    $g \in G \quad$  $F(g * \mathbf{v}) = F(\mathbf{v})$
    \item \textbf{$G$-Equivariant} if $\forall \mathbf{v}\in V $,  $g\in G\quad$  $F(g * \mathbf{v}) = g * F(\mathbf{v})$
\end{enumerate}

For our task, we are especially interested in $SO(3)$, the group of rotations of $\mathbb{R}^3$. In fact, 3D symmetries are intertwined with the law of physics describing quantum interactions, i.e. the Schrodinger equations \citep{schrodinger1926an}. Specifically, if we rotate a system the energy should not vary, $SO(3)$-invariance, and forces should rotate accordingly, $SO(3)$-equivariance.

While equivariance is desired in $\mathbb{R}^3$, equivariant models \citep{weiler20183d, goodman2000representations, deng2021vector, puny2022frame} use atom embeddings in spaces of much higher dimension. In particular, they use irreps corresponding to the coefficients of the real spherical harmonics, which represent a function on a sphere. Given a set of coefficients $\mathbf{x}$, we define a spherical function $F_{\mathbf{x}}: S^2 \rightarrow \mathbb{R}$ as:
\begin{equation} \label{eqn:spherical}
    F_{\mathbf{x}}(\hat{\mathbf{r}}) = \sum_{l,m} \mathbf{x}^{(l)}_m \mathbf{Y}_{m}^{(l)}(\hat{\mathbf{r}})
\end{equation}
where $\hat{\mathbf{r}} = \mathbf{r} / |\mathbf{r}|$ is a unit vector indicating the orientation and $\mathbf{Y}_{m}^{(l)}$ are the real spherical harmonic basis functions defined over degrees $l\in[0,L]$ and orders $m \in [-l,l]$.

Spherical harmonics have the special property that they are steerable, i.e., there exists a Wigner D-matrix $\mathbf{D}^{(l)}(\mathbf{R})$ of size $(2l + 1) \times (2l + 1)$ for 3D rotation $\mathbf{R}$ for which:
\begin{equation} \label{eqn:wig_D}
\mathbf{x}^{(l)}\cdot \mathbf{Y}^{(l)}(\mathbf{R}\cdot \hat{\mathbf{r}}) =  (\mathbf{D}^{(l)}(\mathbf{R})\mathbf{x}^{(l)}) \cdot \mathbf{Y}^{(l)}(\hat{\mathbf{r}})
\end{equation}
This is the fundamental property for extending the concept of $SO(3)$ group action to the spherical harmonics' coefficients:
\begin{equation}
    \mathbf{R} * \mathbf{x}^{(l)} = \mathbf{D}^{(l)}(\mathbf{R}) \cdot \mathbf{x}^{(l)}
\end{equation}
\subsection{Machine learning potentials}

Classically, the prediction of a molecule's energy and forces \citep{behler2016perspective} using machine learning relied on hand-crafted representations \citep{behler2016perspective} like MMFF94 \citep{halgren1996merck} and sGDML \cite{chmiela2018towards}. The research on machine learning potentials has recently moved towards end-to-end learnable models based on graph neural networks \citep{kipf2017semisupervised, gori2005new}. 

Early work on GNNs focuses on models that extract scalar representations from the atoms' positions. They achieve equivariance to rotations by utilizing only invariant features. CGCNN \citep{xie2018crystal} and SchNet \citep{schutt2018schnet} use pair-wise distances, DimeNet \citep{klicpera2020directional}, SphereNet \citep{liu2022spherical} and GemNet \citep{gasteiger2021gemnet, gasteiger2022gemnet} extend this to explicitly capture triplet and quadruplet angles. 

More recently, equivariant models \cite{batzner20223} have surpassed invariant GNNs on small molecular datasets including MD17 \citep{chmiela2017machine} and QM9 \citep{ramakrishnan2014quantum}.
These models build upon the concepts of steerability and equivariance introduced by Cohen and Welling \citep{cohen2016group}. 
They use geometric tensors as node embeddings and ensure equivariance to $SO(3)$ by placing constraints on the operations that can be performed \citep{kondor2018clebsch, fuchs2020se}. 
Specifically, they compute linear operations with a generalized tensor product between the atom embeddings and edges' directions.
In particular, Tensor Field Networks \citep{thomas2018tensor}, NequIP \citep{batzner20223}, SEGNN \citep{brandstetter2022geometric}, MACE \citep{batatia2022mace}, Allegro \citep{musaelian2023learning} and Equiformer \citep{liao2023equiformer} lie in this category, and they are commonly referred to as \textit{e3nn networks}.
Most recently, SCN \citep{zitnick2022spherical} have surpassed invariant GNNs on the large-scale OC-20 \citep{OC20} dataset. While it represents atoms' embeddings by geometric tensors, SCN doesn't strictly enforce equivariance and doesn't compute any tensor product. 

\subsection{e3nn networks}

In e3nn networks \citep{geiger2022e3nn}, an atom $i$'s embedding is encoded using the spherical harmonic coefficients $(\mathbf{x}_{ic})_{m}^{(l)}$ where $0 \le l \le L$, $-l \le m \le l$ and $1 \le c \le C_{l}$ is the number of channels. Assuming a fixed number of channels $C$ over different degrees of the spherical harmonics, the size of the embedding for each atom is $(L+1)^2\times C$. These embeddings are referred to as irreps, since they rotate following the Wigner D-matrices which are the irreducible representations of $SO(3)$ \citep{goodman2000representations}.

The proper use of the irreps constrains the networks' operations to be $SO(3)$-equivariant. The building block of the message passing between two neighboring atoms $s$ and $t$ is the result $\mathbf{a}_{st}$ of an equivariant convolution, which are summed across neighboring atoms $\mathcal{N}_t$ to obtain the updated embedding $\mathbf{x}'_t$:
\begin{equation}\label{eqn:message}
\mathbf{x}'_t = \frac{1}{|\mathcal{N}_t|}\sum_{s \in \mathcal{N}_t}\mathbf{a}_{st}
\end{equation}
The convolutions are obtained by computing a generalized tensor product $\otimes^{l_o}_{l_i, l_f}$ between the input embedding $\mathbf{x}_t$ and the edge direction (filter), $\hat{\mathbf{r}}_{st} = \mathbf{r}_{st} / |\mathbf{r}_{st}|$.  The values of $\mathbf{a}_{st}$ of degree $l_o$ are obtained as a sum of irreps:
\begin{equation}\label{eqn:equivconv}
\mathbf{a}_{st}^{(l_o)} = \sum_{l_i, l_f} \mathbf{x}_s^{(l_i)} \otimes^{l_o}_{l_i, l_f} \mathbf{h}_{l_i, l_f, l_o}  \mathbf{Y}^{(l_f)}(\hat{\mathbf{r}}_{st})
\end{equation}

$\mathbf{h}_{l_i, l_f, l_o} := \mathbf{F}_{l_i, l_f, l_o}(|\mathbf{r}_{st}|, z_s, z_t)$ is a learnable non-linear function that takes as input the distance between the atoms and their atomic numbers $z_s$ and $z_t$, and outputs a scalar coefficient.

For fixed integer values of $l_i$, $l_f$ and $l_o$, the generalized tensor product $\otimes^{l_o}_{l_i, l_f}$ is a bilinear equivariant operation that takes as input an irrep of degree $l_i$ and a filter irrep of degree $l_f$ and outputs an irrep of degree $l_o$. We can compute the tensor product of Equation (\ref{eqn:equivconv}) using the Clebsch-Gordan coefficients $\mathbf{C}^{(l_o, m_o)}_{(l_i, m_i), (l_f, m_f)}$ \citep{griffiths2018introduction} as follows:
\begin{equation}\label{eqn:C-G}
\begin{gathered}
    \left(\mathbf{x}_s^{(l_i)} \otimes^{l_o}_{l_i, l_f} \mathbf{h}_{l_i, l_f, l_o}  \mathbf{Y}^{(l_f)}(\hat{\mathbf{r}}_{st})\right)_{m_o}^{(l_o)} = \\ \sum_{m_i, m_f} (\mathbf{x}_s^{(l_i)})_{m_i}  \mathbf{C}^{(l_o, m_o)}_{(l_i, m_i), (l_f, m_f)}  \mathbf{h}_{l_i, l_f, l_o}  \mathbf{Y}^{(l_f)}_{m_f}(\hat{\mathbf{r}}_{st})
\end{gathered}
\end{equation}
The values of $l_i, l_f$ and $l_o$ of the tensor product in Equation (\ref{eqn:equivconv}) are specified in the architecture of the model and are non-zero for $|l_o - l_i | \le l_f \le l_i + l_o$.
Retaining all the non-zero tensor products $\otimes^{l_o}_{l_i, l_f}$ up to degree $L$ becomes computationally unfeasible as we scale $L$ since it requires $O(L^3 \cdot C)$ 3D matrix multiplications. Therefore most e3nn networks are limited to the computation of $L=1$ or $2$ and $l_f \le 2$ or $3$.

\section{Efficient equivariant convolution}\label{sec:eq_eff}
\begin{figure}
  \centering
    \includegraphics[width=.23\textwidth]{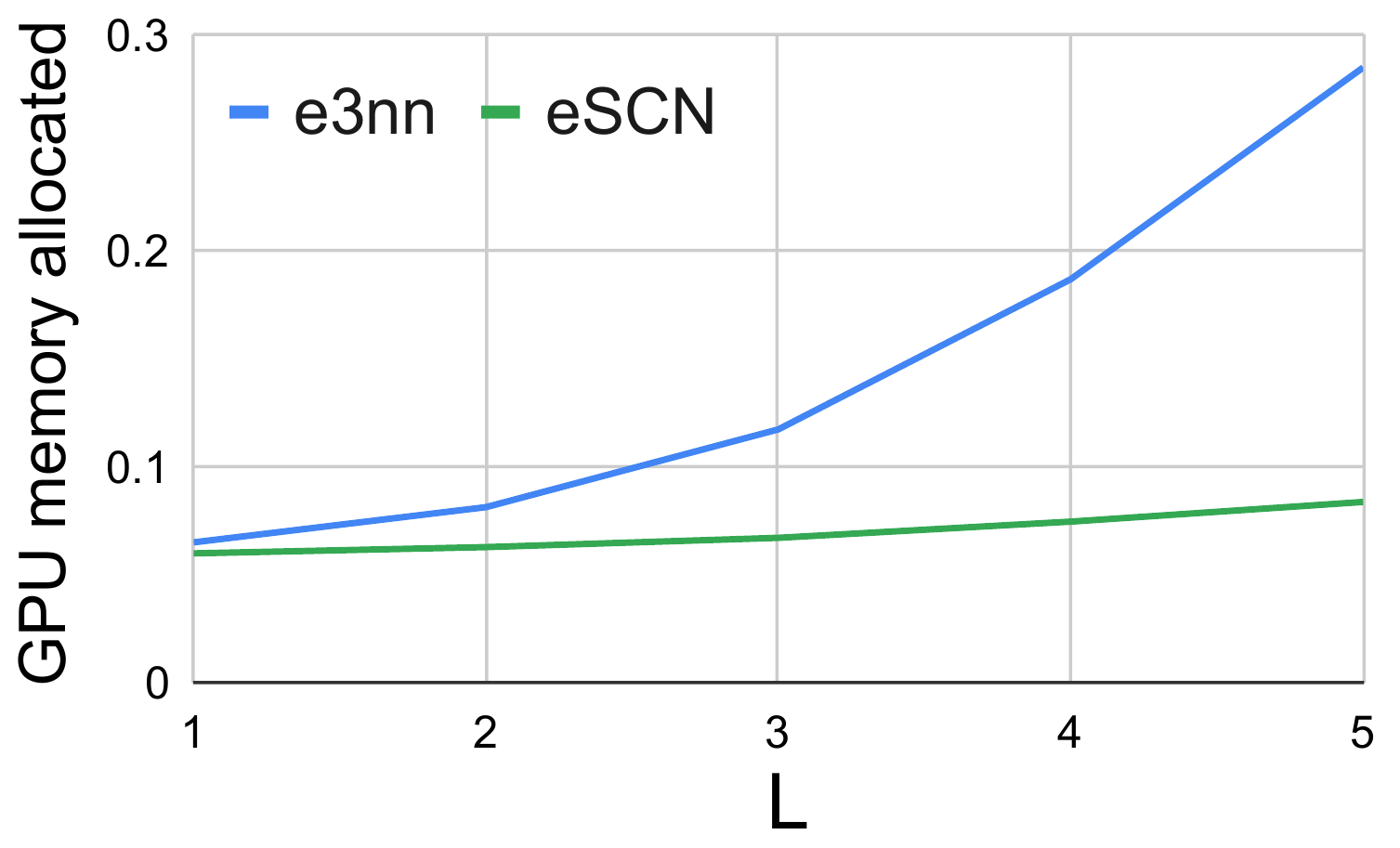}
\includegraphics[width=.23\textwidth]{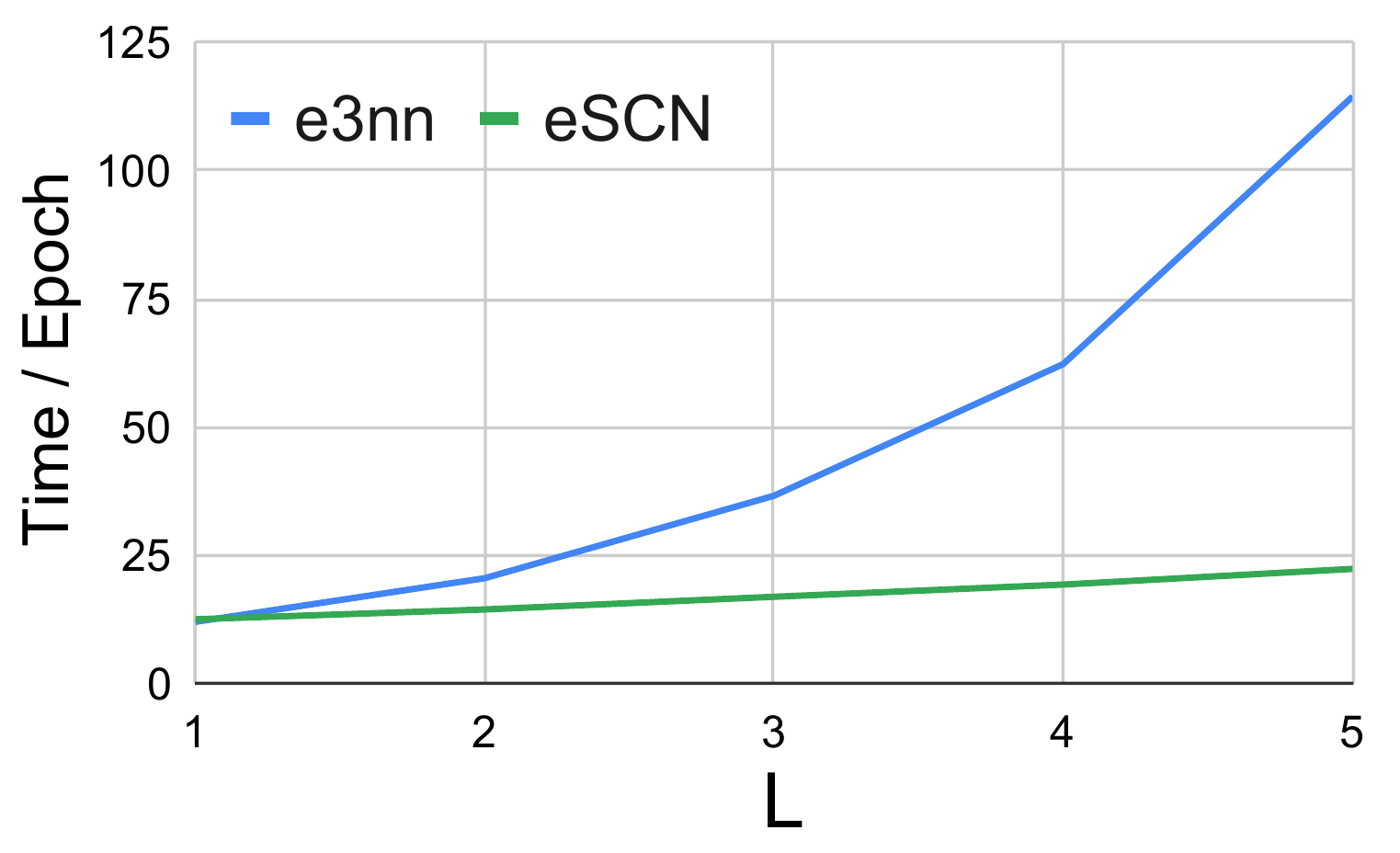}
    \vspace{-6pt}
  \caption{We implement two mathematically equivalent equivariant GNNs. \textit{e3nn} uses the e3nn PyTorch library, whereas \model~uses our novel approach.
\textbf{Left:} GPU memory allocated (\%) at training time with resepct to the maximum degree of the spherical harmonics ($L$). \textbf{Right:} time (h) per epoch by $L$. We fix the number of channels to $C=64$ and remove any non-linearity.}
\label{fig:e3nn_comp}
\end{figure}
The computational complexity of performing a full tensor product makes it extremely expensive to use degrees higher than $L=1$ or $2$. For many applications, such as modeling systems of atoms, high-fidelity of angular information is critical for modeling their interactions. Next, we demonstrate how to dramatically speed up these calculations for higher degrees.

\begin{figure*}
  \centering
    \includegraphics[width=\textwidth]{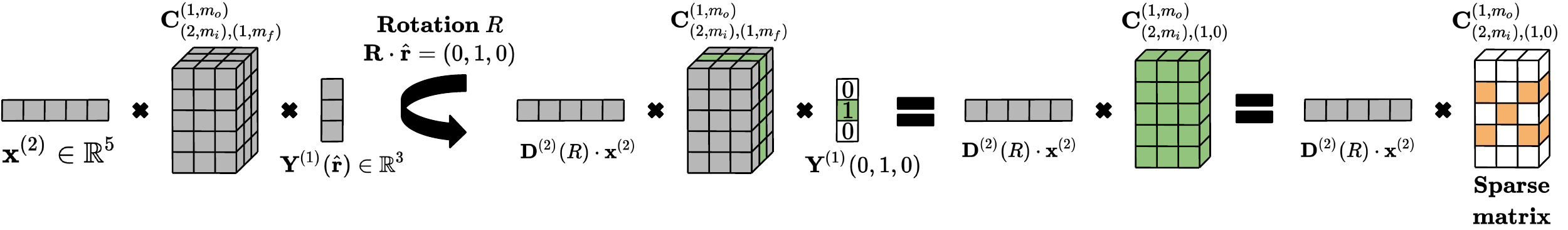}
    \vspace{-6pt}
  \caption{
  Visual representation of the tensor product $\mathbf{x}^{(2)} \otimes^{1}_{2, 1} \mathbf{Y}^{(l)}(\hat{\mathbf{r}})$ between $\mathbf{x} \in \mathbb{R}^5$ and $\mathbf{Y}^{(1)}(\hat{\mathbf{r}}) \in \mathbb{R}^3$. The computational cost of the tensor product can be lowered if the direction $\hat{\mathbf{r}}$ is aligned with the y-axis. The tensor product is then reduced to the multiplication of the green shaded entries. The orange shaded entries are the non-zero entries in the Clebsch-Gordan matrix.
  }
  \label{fig:C-G_1}
\end{figure*}

Let $F$ be an $SO(3)$-equivariant function, then the following holds for any Wigner-D rotation matrix $\mathbf{D}^{(l)}$, rotation matrix $\mathbf{R}$ and $\mathbf{x}^{(l)}$ irreps of degree $l$:
\begin{equation}
F(\mathbf{x}^{(l)}) = \mathbf{D}^{(l)}(\mathbf{R}^{-1})\cdot F(\mathbf{D}^{(l)}(\mathbf{R}) \cdot \mathbf{x}^{(l)})
\label{eqn:equiv}
\end{equation}
Combining Equations (\ref{eqn:wig_D}), (\ref{eqn:equiv}) with (\ref{eqn:equivconv}) we find:
\begin{equation}
\mathbf{a}_{st}^{(l_o)} = \mathbf{D}^{-1} \cdot \sum_{l_i, l_f} \tilde{\mathbf{x}}_s^{(l_i)} \otimes^{l_o}_{l_i, l_f} \mathbf{h}_{l_i, l_f, l_o}  \mathbf{Y}^{(l_f)}(\mathbf{R} \cdot \hat{\mathbf{r}}_{st})
\label{eqn:equivconvsub}
\end{equation}
where  $\tilde{\mathbf{x}}_s^{(l_i)} = \mathbf{D}^{(l_i)}(\mathbf{R})\cdot \mathbf{x}_s^{(l_i)}$ and $\mathbf{D}^{-1} = \mathbf{D}^{(l_o)}(\mathbf{R}^{-1})$. 

We can now present our key observation. By choosing a specific $\mathbf{R}$, we can reduce the cost of computing Equation (\ref{eqn:equivconvsub}) substantially. Specifically, if we select a rotation matrix $\mathbf{R}_{st}$ so that $\mathbf{R}_{st} \cdot \hat{\mathbf{r}}_{st} = (0, 1, 0)$, we find $\mathbf{Y}(\mathbf{R}_{st} \cdot \hat{\mathbf{r}}_{st})$ becomes sparse:
\begin{equation}
    \mathbf{Y}^{(l)}_{m}(\mathbf{R}_{st} \cdot \hat{\mathbf{r}}_{st}) \propto \mathbf{\delta}^{(l)}_m = \begin{cases}
     1 & \text{if } m = 0 
    \\
     0  & \text{if } m \neq 0
\end{cases}
\label{eqn:delta}
\end{equation}
Note that this observation is widely used in computational electromagnetics; specifically for the fast multipole method, this is known as the point-and-shoot method for calculating translation operators \citep{wala2020optimization}.

Substituting Equation (\ref{eqn:delta}) into the right-hand side of Equation (\ref{eqn:equivconvsub}) results in the following:
\begin{equation}
\mathbf{a}_{st}^{(l_o)} = \mathbf{D}_{st}^{-1}\cdot \sum_{l_i, l_f} \tilde{\mathbf{x}}_s^{(l_i)} \otimes^{l_o}_{l_i, l_f} \mathbf{h}_{l_i, l_f, l_o}  \mathbf{\delta}^{(l_f)}
\label{eqn:equivconvsub2}
\end{equation}
This reduces significantly the amount of computation for the tensor product $\otimes^{l_o}_{l_i, l_f}$ since we don't need to sum over $m_f$ in Equation (\ref{eqn:C-G}):
\begin{equation}
\begin{gathered} \label{eqn:C-G_red}
    \left(\tilde{\mathbf{x}}_s^{(l_i)} \otimes^{l_o}_{l_i, l_f} \mathbf{h}_{l_i, l_f, l_o} \delta^{(l_f)}\right)_{m_o}^{(l_o)} = \\
     \sum_{m_i}  (\tilde{\mathbf{x}}_s^{(l_i)})_{m_i}  \mathbf{C}^{(l_o, m_o)}_{(l_i, m_i), (l_f, 0)}\mathbf{h}_{l_i, l_f, l_o} 
\end{gathered}
\end{equation}
However, even further computational efficiency gains may be achieved. The Clebsch-Gordan matrix of Equation (\ref{eqn:C-G_red}) is sparse and follows a particular pattern; see Appendix \ref{proof:C-G} for the proof.
\begin{proposition}\label{prop:C-G}
    The coefficient $\mathbf{C}^{(l_o, m_o)}_{(l_i, m_i), (l_f, 0)}$ is non-zero only if $m_i = \pm m_o$.
    Moreover $\mathbf{C}^{(l_o, m)}_{(l_i, m), (l_f, 0)} = \mathbf{C}^{(l_o, -m)}_{(l_i, -m), (l_f, 0)}$ and $\mathbf{C}^{(l_o, m)}_{(l_i, -m), (l_f, 0)} = - \mathbf{C}^{(l_o, -m)}_{(l_i, m), (l_f, 0)}$      
\end{proposition}
\begin{figure}
  \centering
    \includegraphics[width=0.45\textwidth]{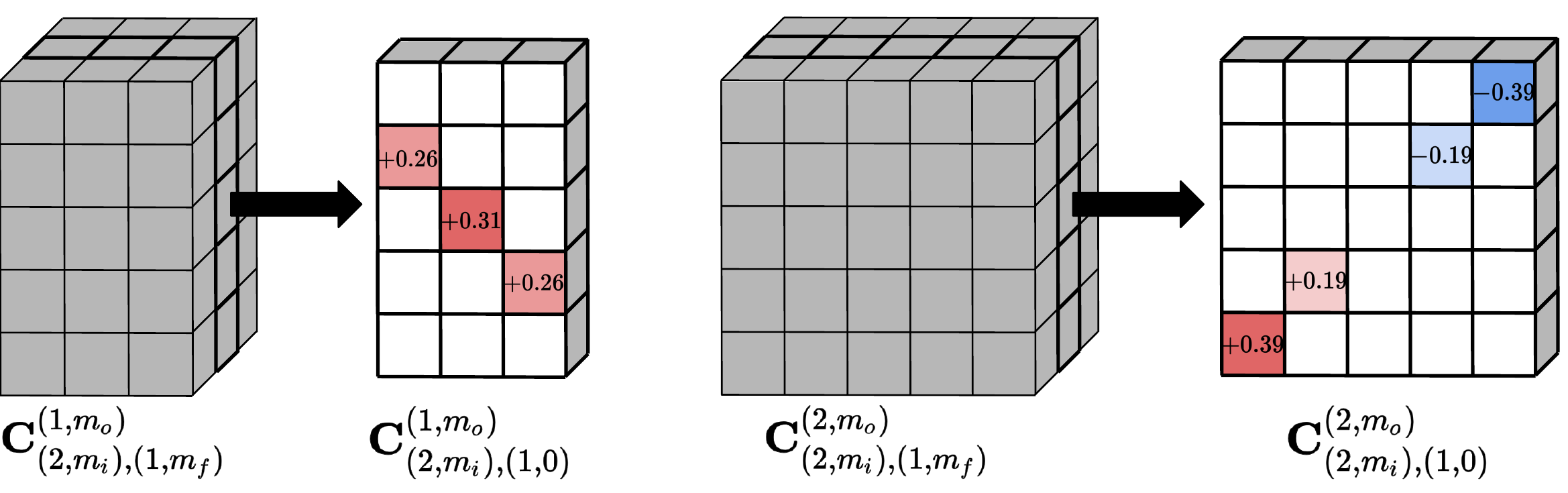}
\caption{Visual representation of the Clebsch-Gordan matrices $\mathbf{C}^{(1, m_o)}_{(2, m_i), (1, m_f)} \in \mathbb{R}^{5 \times 3 \times 3}$ and $\mathbf{C}^{(2, m_o)}_{(2, m_i), (1, m_f)} \in \mathbb{R}^{5 \times 3 \times 5}$. We illustrate the sparsity of the Clebsch-Gordan matrices as stated in Proposition \ref{prop:C-G}.}
\end{figure}

Therefore the Clebsch-Gordan coefficients can be compactly described using the following notation:
\begin{equation}\label{eqn:C-G_not}
(\mathbf{c}_{l_i,l_f,l_o})_{m} = \begin{cases}
     \mathbf{C}^{(l_o, m)}_{(l_i, m), (l_f, 0)}  & \text{if } m > 0 
     \\
     \mathbf{C}^{(l_o, 0)}_{(l_i, 0), (l_f, 0)}  & \text{if } m = 0 
     \\
     \mathbf{C}^{(l_o, -m)}_{(l_i, m), (l_f, 0)}  & \text{if } m < 0
\end{cases}
\end{equation}
We further simplify the right-hand side of Equation (\ref{eqn:equivconvsub2}) using the result in Proposition \ref{prop:C-G} and the notations introduced above obtaining:
\begin{equation}
    \mathbf{a}_{st}^{(l_o)} = \mathbf{D}_{st}^{-1} \cdot \sum_{l_i} \mathbf{w}^{(l_o)}_{st, l_i}
\end{equation}
where $\mathbf{w}^{(l_o)}_{st, l_i}$ is defined by the following Equations:
\begin{equation}
\begin{split}
&\textit{for } m_o  > 0:  \\
&\begin{pmatrix}
(\mathbf{w}^{(l_o)}_{st, l_i})_{m_o}  \\
(\mathbf{w}^{(l_o)}_{st, l_i})_{-m_o} 
\end{pmatrix}  = \begin{pmatrix}
\tilde{\mathbf{h}}^{(l', l)}_{m_o} & -\tilde{\mathbf{h}}^{(l', l)}_{-m_o} \\
\tilde{\mathbf{h}}^{(l', l)}_{-m_o} & \tilde{\mathbf{h}}^{(l', l)}_{m_o}
\end{pmatrix} \cdot \begin{pmatrix}
(\tilde{\mathbf{x}}^{(l_i)}_{s})_{m_o}  \\
(\tilde{\mathbf{x}}^{(l_i)}_{s})_{-m_o} 
\end{pmatrix} \\
&\textit{for }  m_o  = 0:  \\
&\quad \quad \quad \quad \quad \quad (\mathbf{w}^{(l_o)}_{st, l_i})_{0} = \tilde{\mathbf{h}}^{(l_i, l_o)}_{0}  (\tilde{\mathbf{x}}_s^{(l_i)})_{0} 
\label{eqn:SO2_conv}
\end{split}
\end{equation}
and $\tilde{\mathbf{h}}^{(l_i, l_o)}$ satisfies:
\begin{equation}
\tilde{\mathbf{h}}^{(l_i, l_o)}_{m} = \sum_{l_f}  \left( \mathbf{h}_{l_i, l_f, l_o} (\mathbf{c}_{l_i,l_f,l_o})_m \right) 
\label{eqn:h_new}
\end{equation}
We prove in Appendix \ref{app:bij} that the scalar coefficients $\tilde{\mathbf{h}}^{(l_i, l_o)}_{m}$ are related to $\mathbf{h}_{l_i, l_f, l_o}$ with a linear bijection. Therefore the model can directly parametrize $\tilde{\mathbf{h}}^{(l_i, l_o)}_{m}$ without loosing or adding any information.

The new formulation of the equivariant convolution provided in Equation (\ref{eqn:SO2_conv}) doesn't pre-compute the Clebsch-Gordan coefficients and is drastically more efficient since we no longer sum over $m_f$, $m_i$ and $l_f$. 

In Appendix \ref{app:comp}, we provide a theoretical analysis on the computational complexity of the equivariant convolutions; specifically, we show that the computational cost of performing a full tensor product scales with $O(L^6)$, whereas our efficient approach is $O(L^3)$.
Moreover we show how the new formulation of the equivariant convolutions can be efficiently parallelized on a GPU by reshaping the irreps to group coefficients of the same order \textit{m} together. This allows to perform a separate convolution for each order $\pm$$\textit{m}$ by parametrizing only two dense linear transformations. Note that for higher \textit{m} close to $L$, the GPU may not be fully utilized as the number of coefficients decreases, but this doesn’t happen in practice since we typically restrict to $|\textit{m}| \le $ $2$ or $3$.

\section{SO(2) formulation}

In the previous section, we described how the computation of the generalized tensor product can be greatly simplified. Here we propose an alternative explanation of the same operation as a reduction from $SO(3)$-equivariance to $SO(2)$-equivariance. Previously, we simplified the generalized tensor product by first rotating the spherical harmonic coefficients so that the primary axis (y-axis) was aligned with the direction of messages's source atom $s$ to the target atom $t$, $\hat{\mathbf{r}}_{st}$. Intuitively, once this axis is fixed, only a single rotational degree of freedom remains; the roll rotation over $\hat{\mathbf{r}}_{st}$. Thus, if the message passing function is $SO(2)$-equivariant to the roll rotation about $\hat{\mathbf{r}}_{st}$, it will also be $SO(3)$-equivariant. 

To see this mathematically, let us define the circular harmonics, which are the $SO(2)$ representation analogous to spherical harmonics in $SO(3)$.  A circular harmonic of degree $k$ and order $j$ is $\mathbf{B}^{(k)}_j : S \rightarrow \mathbb{R}$ defined as:
\begin{equation} \label{eqn:cirhar}
    \mathbf{B}^{(k)}_j(\phi) = \begin{cases}
        sin(k \phi ) & \textit{ if } \quad j =1 \\
        cos(k \phi ) & \textit{ if } \quad j =-1 \\
    \end{cases}
\end{equation}
\begin{figure}
  \centering
    \includegraphics[width=0.4\textwidth]{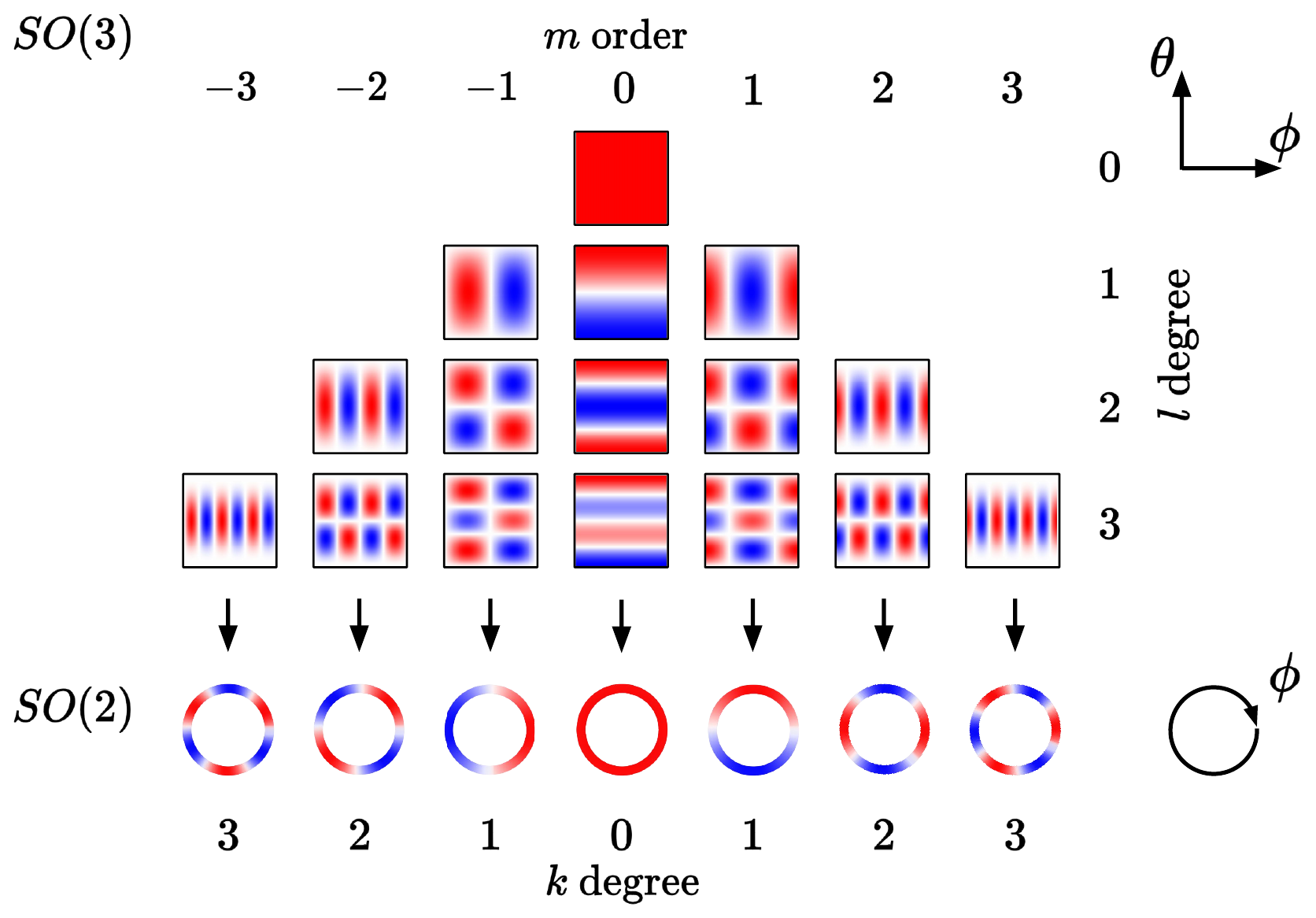}
  \caption{Illustration of how the spherical harmonics (top) are reduced to a set of circular harmonics (bottom) when $\theta$ is held constant.}
  \label{fig:proj}
\end{figure}
The real spherical harmonics basis functions used in Equation (\ref{eqn:spherical}) can be written as:
\begin{equation} \label{eqn:sphhar}
\mathbf{Y}_{m}^{(l)}(\theta, \phi) = 
\begin{cases}
     P_m^{(l)}(\theta)\sin(m\phi) & \text{if } m > 0 
     \\
      P_m^{(l)}(\theta)\cos(m\phi)  & \text{if } m \leq 0
\end{cases}
\end{equation}
where $P_m^{(l)}$ is the Legendre polynomial which depends only on the colatitude angle $\theta \in [ 0, \pi ]$ and $\phi \in [0, 2\pi]$ is the longitude angle, i.e., the rotation about the y-axis. Note that $\theta$ and $\phi$ may be computed directly from $\hat{\mathbf{r}}$. Combining Equations (\ref{eqn:spherical}), (\ref{eqn:cirhar}) and (\ref{eqn:sphhar}) we find the spherical function $F_{\mathbf{x}}$ can be rewritten as:
\begin{equation} \label{eqn:combinehar}
F_{\mathbf{x}}(\theta,\phi) = \sum_{l,m} \mathbf{x}^{(l)}_m P_m^{(l)}(\theta) \mathbf{B}^{(m)}_{sgn(m)}(\phi), 
\end{equation}
where $sgn(m) = -1$ if $m \leq 0$ and 1 otherwise. Using our proposed approach, if atoms are rotated in 3D, only the angle $\phi$ will change since $\theta$ is fixed by aligning $\hat{\mathbf{r}}_{st}$ to the y-axis. 
As a result, Equation (\ref{eqn:combinehar}) is equivalent to a sum of 2D circular harmonics and $\mathbf{x}^{(l)}_m$ becomes a coefficient of $\mathbf{B}^{(m)}_{sgn(m)}$. Additionally, the circular harmonics themselves are a Fourier series; as such, a convolution about $\phi$, which is $SO(2)$ equivariant \citep{worrall2017harmonic}, may be performed using a simple point-wise product in the spectral domain \citep{maxime1906introduction}:
\begin{equation}
\begin{pmatrix}
(\mathbf{x}')^{(l)}_{m}  \\
(\mathbf{x}')^{(l)}_{-m}
\end{pmatrix} := \begin{pmatrix}
\mathbf{y}_{l, m} & -\mathbf{y}_{l, -m} \\
\mathbf{y}_{l, -m} & \mathbf{y}_{l, m}
\end{pmatrix} \cdot
\begin{pmatrix}
\mathbf{x}^{(l)}_{m}  \\
\mathbf{x}^{(l)}_{-m} 
\end{pmatrix}
\end{equation}
Furthermore, we can generalize the convolution about $\phi$ adding self-interactions between different circular channels as follows:
\begin{equation}
\begin{pmatrix}
(\mathbf{x}')^{(l)}_{m}  \\
(\mathbf{x}')^{(l)}_{-m}
\end{pmatrix} := \sum_{l'}\begin{pmatrix}
\mathbf{y}_{l, m, l'} & -\mathbf{y}_{l, -m, l'} \\
\mathbf{y}_{l, -m, l'} & \mathbf{y}_{l, m, l'}
\end{pmatrix} \cdot
\begin{pmatrix}
\mathbf{x}^{(l')}_{m}  \\
\mathbf{x}^{(l')}_{-m} 
\end{pmatrix}
\end{equation}
We finally note that the above Equation coincides with tensor product of Equation (\ref{eqn:SO2_conv}) when we let $ \mathbf{y}_{l, m, l'} = \tilde{\mathbf{h}}_{m}^{(l', l)}$. In conclusion, the generalized tensor product of Equation (\ref{eqn:message}) may be viewed as a set of generalized convolutions in $SO(2)$ if the spherical harmonic coefficients are rotated appropriately.

\section{Architecture}
\begin{figure}
  \centering
    \includegraphics[width=0.2\textwidth]{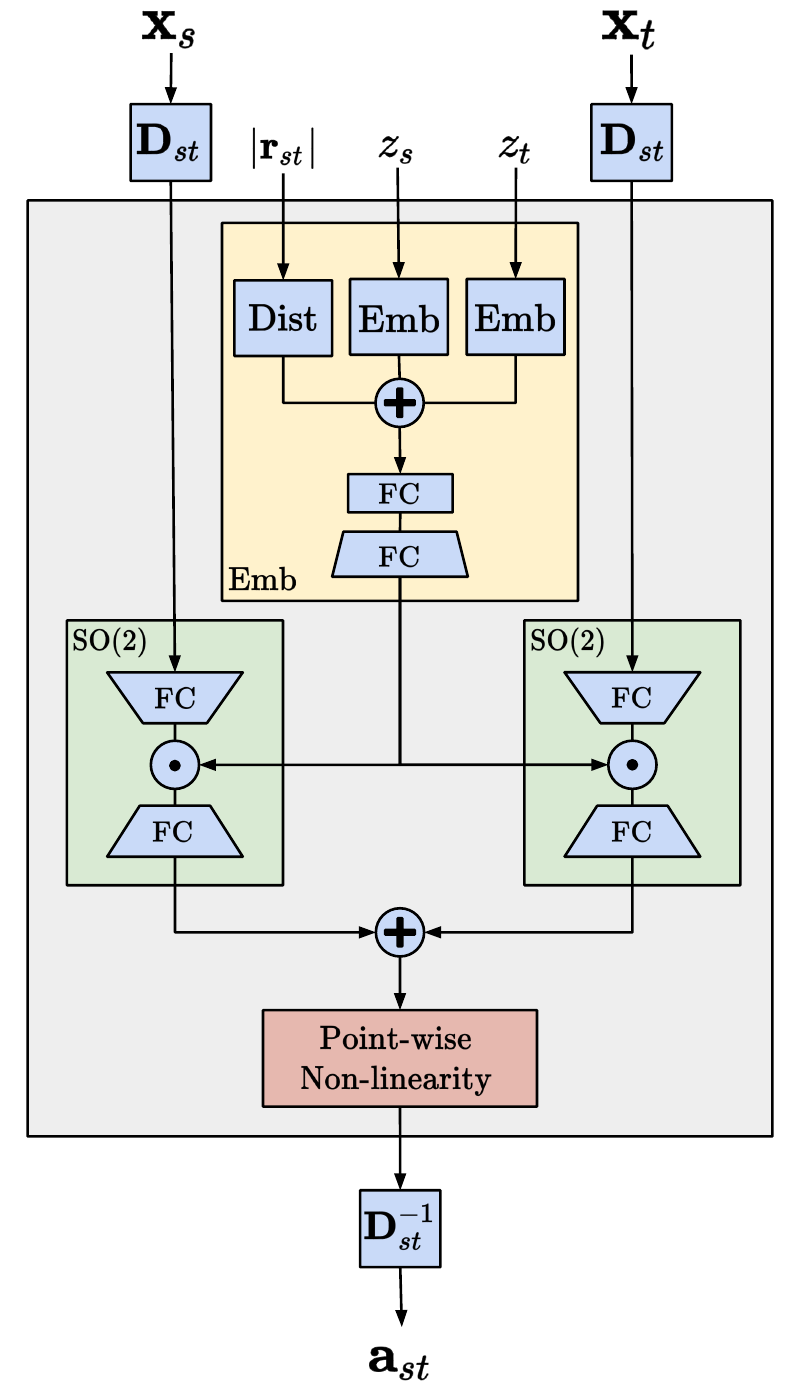}
  \caption{Block diagram of the message passing architecture containing the edge embedding block (yellow), the $SO(2)$ blocks (green) and the point-wise non-linearity (red).}
  \label{fig:method1}
\end{figure}

Given an atomic structure with $n$ atoms, our goal is to predict the structure’s energy $E$ and the per-atom forces $\mathbf{f}_i$ for each atom $i$. These values are estimated using a Graph Neural Network (GNN) where each node represents an atom and edges represent nearby atoms. As input, the network is given the vector distance $\mathbf{r}_{ij}$ between atom $i$ and atom $j$, and each atom’s atomic number $z_i$. The neighbors $\mathcal{N}_i$ for an atom $i$ are determined using a fixed distance threshold with a maximum number of neighbors set to 20.

Each node $i$'s embedding is a set of irreps $\mathbf{x}_i$, where $\mathbf{x}_i$ is indexed by $l$, $m$ and $c$. $\mathbf{x}_{imc}^{(l)}$ is the $m$-th component (order) of the $c$-th channel of an irrep of degree $l$. 

$\mathbf{x}_{i}^{(0)}$ is initialized with an embedding based on the atomic number $z_i$ and $\mathbf{x}_{i}^{(l)}$ are set to $0$ for $l \ne 0$. 
The nodes' embeddings $\mathbf{x}_{i}$ are updated by the GNN through message passing for $k \in K$ layers to obtain the final node embeddings. The nodes’ embeddings are updated by first calculating a set of messages $\mathbf{a}_{st}$ for each edge, which are then aggregated at each node. Finally, the energy and forces are estimated from the nodes' final embeddings.

Our model follows the architecture of SCN \cite{zitnick2022spherical}, except we replace the message passing operation with our equivariant approach. The message aggregation and output blocks remain the same as SCN. While not novel, we describe for completeness the message aggregation and output blocks in Sections \ref{sec:aggr} and \ref{sec:output}. 

\subsection{Message passing}
\begin{figure*}[htb]
\begin{center}
\includegraphics[width=.74\textwidth]{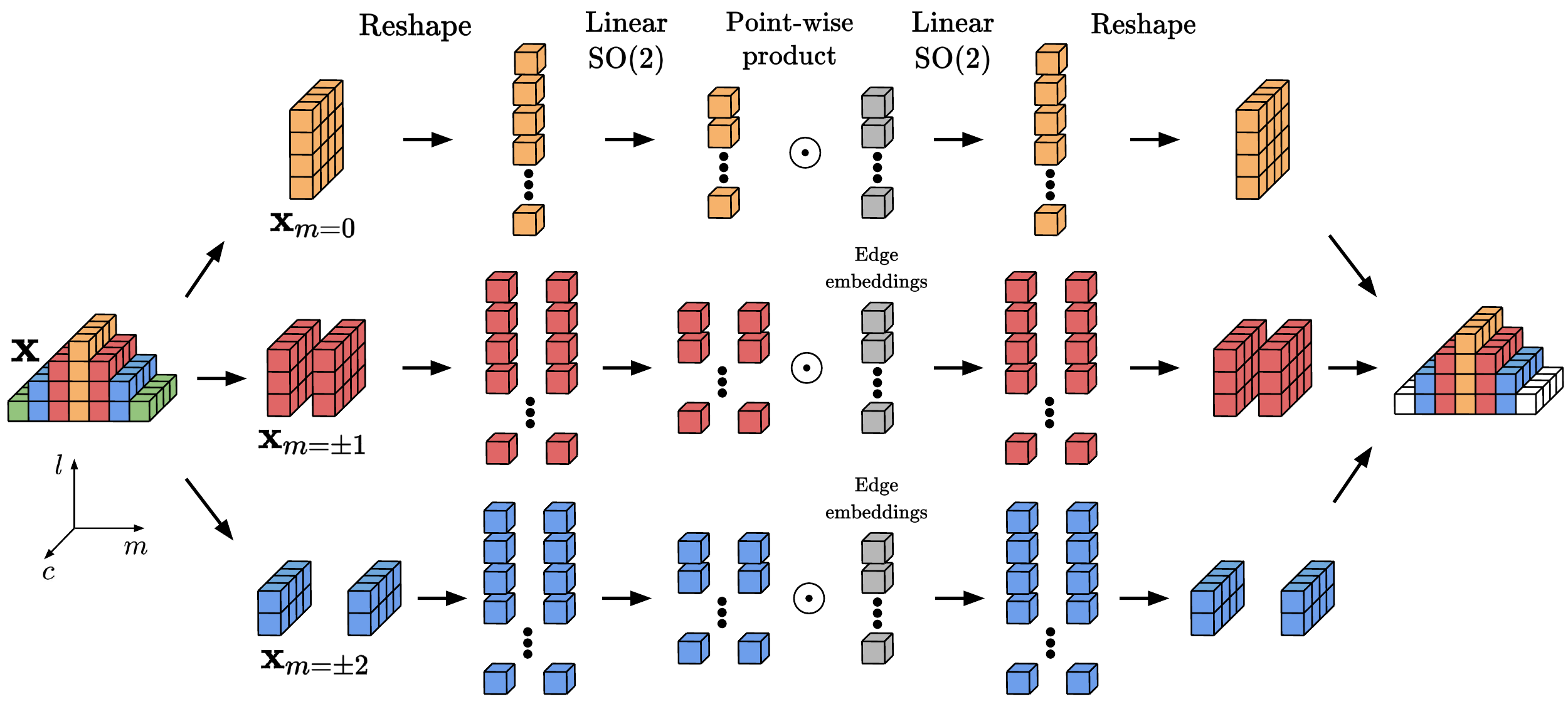}
\caption{The $SO(2)$ block for $L=4$, $C=4$ and $M = 2$. The orange, red, blue and green shaded entries are the coefficients of the spherical harmonics of $m=0, 1, 2$ and $3$ respectively.} 
\label{fig:method2}
\end{center}

\end{figure*}
Given a target node $t$ and its neighbors $s \in \mathcal{N}_t$, we want to update the embeddings $\mathbf{x}_t$ at each layer $k$. The message passing architecture can be divided into three different parts: 1) An edge embedding block that computes an embedding from invariant edge information, 2) a block that performs the generalized convolution in $SO(2)$, 3) a point-wise spherical non-linearity performed on the sphere. In Figure \ref{fig:method1} and \ref{fig:method2} we illustrate the message passing architecture and a tensor representation of the $SO(2)$ block. 

\textbf{Edge embedding block.} 
It takes as input the distance $|\mathbf{r}_{st}|$ and the atomic numbers $z_s$ and $z_t$, and outputs two invariant embeddings per order $m$ of size $H=256$. The atomic numbers $z_s$ and $z_t$ are used to look up two independent embeddings of size $128$, and a set of 1D basis functions are used to represent $|\mathbf{r}_{st}|$ using equally spaced Gaussians every $0.02 \mathring{A}$ from $0$ to the $12 \mathring{A}$ with $\sigma = 0.04$ followed by a linear layer. Their values are added together and passed through a two-layer neural network with non-linearities to produce the embeddings. They are fed into the $SO(2)$ block.

\textbf{SO(2) block.} 
The model computes two independent equivariant convolutions for $\mathbf{x}_s$ and $\mathbf{x}_t$. As described previously for enabling efficient calculations, the atoms' embeddings are rotated to align the direction $\hat{\mathbf{r}}_{st}$ to the y-axis. The rotations are performed using the Wigner D-matrices $\mathbf{D}(R_{st})$ so that $\mathbf{R}_{st} \cdot \hat{\mathbf{r}}_{st} = (0,1,0)$.
The generalized $SO(2)$ convolutions are performed on all coefficients up to order $M \leq L$.  For each $m \in M$, there are $L + 1 - m$ coefficients with $C$ channels. Instead of doing a large dense linear layer to perform the generalized convolution, we linearly project the coefficients down to a $H=256$ hidden layer. The hidden layer values are multiplied by the invariant edge embeddings to incorporate the edge information and then linearly expanded back to the original layer size of $(L + 1 - m)\times C$. The implementation of the $SO(2)$ convolutions inherently calculates two linear layers, used to update the values of the the $m$ and $-m$ coefficients as described in Equation (\ref{eqn:SO2_conv}).
Note that as we demonstrate later, $M$ may be less than $L$ with minimal loss in accuracy while improving efficiency. The ability to use smaller $M$ may be due to a couple of reasons. First, larger $m$ have fewer coefficients, i.e., the number of coefficients for a specific $m$ is $L - m + 1$. Second, after the coefficients are rotated to align with an edge, larger $m$ encode the higher frequency information perpendicular to the edge's direction, which may be less informative than information parallel to the edge's direction.

\textbf{Point-wise spherical non-linearity.} The result of the two $SO(2)$ blocks are added together and a point-wise spherical non-linearity is performed:
\begin{equation} \label{eqn:pointwise}
     \mathbf{x} \rightarrow \int \mathbf{Y} (\hat{\mathbf{r}}) \text{SiLU}(F_{\mathbf{x}}(\hat{\mathbf{r}})) \, d\hat{\mathbf{r}}
\end{equation}
A point-wise function is an equivariant operation since it performs the same function at all points on the sphere \citep{cohen2016steerable, cohen2018spherical, poulenard2021functional}. For this operation we use the SiLU activation function \cite{elfwing2018sigmoid} $(\mathbb{R} \rightarrow \mathbb{R})$. In practice a discrete approximation of the integral is performed. See the appendix for a discussion on the sampling strategy. The message $\mathbf{a}_{st}$ is finally obtained by performing a rotation back to the original coordinate frame using the Wigner D-matrix $\mathbf{D}(R_{st}^{-1})$.

\subsection{Message aggregation}
\label{sec:aggr}
 For each atom $t$, the messages are aggregated together by taking their sum, $\mathbf{a}_t = \sum_{s} \mathbf{a}_{st}$. Another point-wise spherical non-linear function is then performed using both $\mathbf{a}_t$ and $\mathbf{x}_{t}$ as introduced by SCN \cite{zitnick2022spherical} to enable more complex non-linear interactions of the messages. The result is added to the original embedding $\mathbf{x}_{t}$ to obtain our final updated embedding $\mathbf{x}_{t}'$:
\begin{equation} \label{eqn:pointwise-aggr}
     \mathbf{x}_{t}' = \mathbf{x}_{t} + \int \mathbf{Y} (\hat{\mathbf{r}}) P_{agg}(F_{\mathbf{a}_t}(\hat{\mathbf{r}}), F_{\mathbf{x}_{t}}(\hat{\mathbf{r}})) \, d\hat{\mathbf{r}}
\end{equation}
 For the point-wise function $P_{agg}$ we use a three layer neural network $(\mathbb{R}^{2C} \rightarrow \mathbb{R}^C)$ with SiLU activation functions. As for Equation (\ref{eqn:pointwise}), the integral in Equation (\ref{eqn:pointwise-aggr}) is performed on a discrete grid as discussed in \cite{zitnick2022spherical}.

\begin{table*}[t]
    \centering
    \renewcommand{\arraystretch}{1}
    \setlength{\tabcolsep}{5pt}
    \begin{adjustbox}{width=1.9\columnwidth ,center}
    \begin{tabular}{lrr|cccc|cc|cc}
       \mc{11}{c}{OC20 Test} \\
      \toprule
         & & & \mc{4}{|c|}{\textbf{S2EF}} & \mc{2}{|c|}{\textbf{IS2RS}} & \mc{2}{|c}{\textbf{IS2RE}} \\
         & & \textbf{Train} & Energy MAE & Force MAE & Force Cos & EFwT & AFbT & ADwT & Energy MAE & EwT \\
        Model & \#Params & \textbf{time} & meV $\downarrow$ & [meV/\AA] $\downarrow$ & $\uparrow$ & [\%] $\uparrow$ & [\%] $\uparrow$ & [\%] $\uparrow$ & meV $\downarrow$ & [\%] $\uparrow$ \\
      %\midrule
        %Median & -- & & 2258 & 84.4 & 0.016 & 0.01 & - & - & - \\
      \midrule
         & & & \mc{7}{c}{\textbf{Train \ocd~All}} \\
        SchNet \cite{schutt2018schnet,OC20}  & 9.1M & 194d &  540 & 54.7 & 0.302 & 0.00 & - & 14.4 & 749 & 3.3  \\
        PaiNN \cite{schutt2021equivariant}
            & 20.1M & 67d & 341 & 33.1 & 0.491 & 0.46 & 11.7 & 48.5 & 471 & - \\
        DimeNet++-L-F+E \cite{klicpera_dimenetpp_2020, OC20} & 10.7M & 1600d & 480 & 31.3 & 0.544 & 0.00 & 21.7 & 51.7 & 559 & 5.0 \\
        SpinConv (direct-forces) \cite{shuaibi2021rotation} & 8.5M & 275d & 336 & 29.7 & 0.539 & 0.45 & 16.7 & 53.6 & 437 & 7.8 \\
        GemNet-dT \cite{gasteiger2021gemnet}
            & 32M & 492d & 292 & 24.2 & 0.616 & 1.20 & 27.6 & 58.7 & 400 & 9.9 \\
        GemNet-OC \cite{gasteiger2022gemnet}
            & 39M & 336d & \bf{233} & 20.7 & 0.666 & 2.50 & 35.3 & 60.3 & 355 & - \\
        SCN $L$=8 $K$=20 \cite{zitnick2022spherical} & 271M & 645d & 244 & 17.7 & 0.687 & 2.59 & 40.3 & \bf{67.1} & \bf{330} & 12.6 \\
        \midrule
        \model~$L$=6 $K$=20 & 200M & 600d & 242 & \bf{17.1} & \bf{0.716} & \bf{3.28} & \bf{48.5} & 65.7 & 341 & \bf{14.3} \\
        \midrule
        & & & \mc{7}{c}{\textbf{Train \ocd~All + MD}} \\
        GemNet-OC-L-E \cite{gasteiger2022gemnet}
            & 56M & 640d & \bf{230} & 21.0 & 0.665 & 2.80 & - & - & - & -\\
        GemNet-OC-L-F \cite{gasteiger2022gemnet}
            & 216M & 765d & 241 & 19.0 & 0.691 & 2.97 & 40.6 & 60.4 & - & -\\
        GemNet-OC-L-F+E \cite{gasteiger2022gemnet}
            & - & - & - & - & - & - & - & - & 348 & 15.0 \\
        {SCN $L$=6 $K$=16 4-tap 2-band \cite{zitnick2022spherical}}  & 168M & 414d & \bf{228} & 17.8 & 0.696 & 2.95 & 43.3 & 64.9 & 328 & 14.2 \\
        {SCN $L$=8 $K$=20 \cite{zitnick2022spherical}}  & 271M & 1280d & 237 & 17.2 & 0.698 & 2.89 & 43.6 & \bf{67.5} & \bf{321} & 14.8 \\
        \midrule
        \model~$L$=6 $K$=20 & 200M & 568d & \bf{228} & \bf{15.6} & \bf{0.728} & \bf{4.11} & \bf{50.3} & 66.7 & \bf{323} & \bf{15.6} \\
        %\model~$L=6$ $K=20$ & 200M & 757d & 235 & \bf{16.5} & \bf{0.724} & \bf{3.40} &  &  & 328 \\
      \bottomrule
    \end{tabular}%
     \end{adjustbox}
    \caption{Comparison of \model~to existing GNN models on the S2EF, IS2RS and IS2RE tasks when
trained on the All or All+MD datasets. Average results across all four test splits are reported. We
mark as bold the best performance and those within a small threshold, e.g., within $0.5$ meV$/\text{\AA}$ MAE, which we found to empirically provide a meaningful performance difference. Training time is in GPU days.}
    \label{tab:comp-all}
\end{table*}
\begin{table*}[t]
    \centering
    \renewcommand{\arraystretch}{1}
    \setlength{\tabcolsep}{3pt}
    \begin{adjustbox}{width=1.9\columnwidth ,center}
    \begin{tabular}{lr|cccc|cccc}
       \mc{10}{c}{OC22 Test} \\
      \toprule
         & & \mc{4}{|c|}{\textbf{S2EF ID}} & \mc{4}{|c|}{\textbf{S2EF OOD}}  \\
         & & Energy MAE & Force MAE & Force Cos & EFwT & Energy MAE & Force MAE & Force Cos & EFwT \\
        Model & \#Params &  meV $\downarrow$ & [eV/\AA] $\downarrow$ & $\uparrow$ & [\%] $\uparrow$ & meV $\downarrow$ & [eV/\AA] $\downarrow$ & $\uparrow$ & [\%] $\uparrow$ \\
      \midrule
        Median & --  & 163.4 & 75 & 0.002 & 0.00 &  160.4 & 73 &  0.002 & 0.00 \\
      \midrule
         & & & \mc{6}{c}{\textbf{Train OC22~ only}} \\
        SchNet \cite{schutt2018schnet,OC20}  & 9.1M &   7,924 & 60.1 &  0.363 & 0.00 &   7,925 & 82.3 & 0.220 & 0.00   \\
        PaiNN \cite{schutt2021equivariant}  & 20.1M &   951 & 44.9 &  0.485 & 0.00 &    2,630 &  58.3 & 0.345 & 0.00   \\
        DimeNet++ \cite{klicpera_dimenetpp_2020, OC20} & 10.7M  & 2,095 & 42.6 & 0.606 & 0.00 &  2,475 & 58.5 & 0.436 & 0.00 \\
        SpinConv (direct-forces) \cite{shuaibi2021rotation} & 8.5M  & 836 & 37.7 &  0.591 & 0.00 & 1,944 &  63.1 & 0.412 & 0.00 \\
        GemNet-dT \cite{gasteiger2021gemnet}
            & 32M  & 939 &  31.6 &  0.665 &  0.00 & 1,271 & 40.5 & 0.530 & 0.00 \\
        GemNet-OC \cite{gasteiger2022gemnet}
            & 39M & 374 & 29.4 &  0.691 & 0.02 & 829 &  39.6 & 0.550 & 0.00 \\
        GemNet-OC \cite{gasteiger2022gemnet} with linear referencing \cite{tran2023open}
            & 39M & 357 & 30.0 &  0.692 & 0.02 & 1,057 &  40.0 &  0.552 & 0.00 \\
        \midrule
        \model~$L$=6 $K$=20 with linear referencing \cite{tran2023open} & 200M & \bf{350} & \bf{23.8} & \bf{0.788} & \bf{0.23} & \bf{789} & \bf{35.7} & \bf{0.637} & 0.00 \\
      \bottomrule
    \end{tabular}%
     \end{adjustbox}
    \caption{Comparison of \model~to existing GNN models on the S2EF task when
trained on OC22 \cite{tran2023open}. We mark as bold the best performance and those within a small threshold, e.g., within $0.5$ meV$/\text{\AA}$ MAE, which we found to empirically provide a meaningful performance difference.}
    \label{tab:comp-oc22}
\end{table*}
\subsection{Output blocks}
\label{sec:output}
Finally, the architecture predicts the per-atom forces $\mathbf{f}_i$ and the total energy $E$ from the final atoms' embeddings $\mathbf{x}_i$ using the same approach as SCN \cite{zitnick2022spherical}. For the energy, we take the integral of a function over the sphere:
\begin{equation} \label{eqn:energy}
     E = \sum_{i} \int P_{energy}(F_{\mathbf{x}_{i}}(\hat{\mathbf{r}})) \, d\hat{\mathbf{r}},
\end{equation}
where $P_{energy}$ is a three layer neural network $(\mathbb{R}^{C} \rightarrow \mathbb{R})$. 

For forces we use a similar equation weighted by the unit vector $\hat{\mathbf{r}}$ to compute the 3D forces for each atom $i$:
\begin{equation} \label{eqn:forces}
     \mathbf{f}_i = \int \hat{\mathbf{r}}P_{force}(F_{\mathbf{x}_{i}}(\hat{\mathbf{r}})) \, d\hat{\mathbf{r}},
\end{equation}
where $P_{force}$ is also a three layer neural network $(\mathbb{R}^{C} \rightarrow \mathbb{R})$. For both Equations (\ref{eqn:energy}) and (\ref{eqn:forces}) we use a discrete approximation of the integral by sampling 128 points on the sphere using weighted spherical Fibonacci point sets \citep{zitnick2022spherical, gonzalez2010measurement}. 

\section{Experiments}

\begin{table*}[t]
    \centering
    \renewcommand{\arraystretch}{1}
    \setlength{\tabcolsep}{5pt}
    \begin{adjustbox}{width=1.9\columnwidth ,center}

    \begin{tabular}{lcccc|cccc|cc}
       \mc{11}{c}{\ocd{} 2M Validation} \\
      \toprule
         & & &  &   & \mc{4}{c|}{\textbf{S2EF}} & \mc{2}{c}{\textbf{ IS2RE}} \\
        & &   & & Samples / & Energy MAE  & Force MAE  & Force Cos  & EFwT & Energy MAE & EwT \\
        \textbf{Model} & & & & GPU sec. &  [meV] $\downarrow$ & [meV/\AA] $\downarrow$ &  $\uparrow$ & [\%] $\uparrow$ &  [meV] $\downarrow$ & [\%] $\uparrow$ \\

      \midrule 
        SchNet \cite{schutt2018schnet} & & & & & 1400 & 78.3 & 0.109 & 0.00 & - & - \\
        DimeNet++ \cite{klicpera_dimenetpp_2020} & & & & & 805 & 65.7 & 0.217 & 0.01 & - & - \\
        SpinConv \cite{shuaibi2021rotation} & & & & & 406 & 36.2 & 0.479 & 0.13 & - & - \\
        GemNet-dT \cite{gasteiger2021gemnet} & & & & 25.8 & 358 & 29.5 & 0.557 & 0.61 & 438 & - \\
        GemNet-OC \cite{gasteiger2022gemnet} & & & & 18.3 & 286 & 25.7 & 0.598 & 1.06 & 407 & - \\
        
        \midrule 
        & \textbf{$L$} & \textbf{\# layers} & \textbf{\# batch} & & & & & \\
        \midrule 
        {SCN 1-tap 1-band \cite{zitnick2022spherical}}  & 6 & 12 & 64 & 7.7 & 299 & 24.3 & 0.605 & 0.98 & - & - \\
        {SCN 4-tap 2-band \cite{zitnick2022spherical}}  & 6 & 12 & 64 & 3.5 & \bf{279} & 22.2 & 0.643 & 1.41 & \bf{371} & \bf{11.0} \\
        {SCN 4-tap 2-band \cite{zitnick2022spherical}}  & 6 & 16 & 64 & 2.3 & \bf{279}  & 21.9 & 0.650 & 1.46 & \bf{373} & \bf{11.0} \\
        \midrule 
        {\model~linear}  & 6 & 12 & 96 & 6.8 & 301 & 21.9  & 0.646 & 1.28  & - & - \\
        \midrule 
        {\model }  & 2 & 12 & 96 & 9.7 & 307 & 26.7 & 0.577 & 0.94 & - & - \\
        {\model }  & 4 & 12 & 96 & 7.8 & 291 & 22.2 & 0.637 & 1.39 & - & - \\
        {\model }  & 6 & 12 & 96 & 6.8 & 294 & 21.3  & 0.653 & 1.45  & \bf{376} & \bf{10.9} \\
        {\model }  & 8 & 12 & 96 & 4.4 & 296 & 21.3 & 0.654 & 1.47 & - & - \\
        \midrule 
        {\model~m=0}  & 6 & 12 & 96 & 13.8 & 316 & 26.5 & 0.584 & 0.88 & - & - \\
        {\model~m=1}  & 6 & 12 & 96 & 8.7 & 309 & 23.4 & 0.626 & 1.13 & - & - \\
        {\model~m=2}  & 6 & 12 & 96 & 6.8 & 294 & 21.3  & 0.653 & 1.45 & \bf{376} & \bf{10.9} \\
        {\model~m=3}  & 6 & 12 & 96 & 4.8 & 295 & 21.2 & 0.656 & 1.38 & - & - \\
        {\model~m=4}  & 6 & 12 & 96 & 3.8 & 298 & 21.2 & 0.657 & 1.41 & - & - \\
        \midrule 
        {\model }  & 6 & 4 & 96 & 14.3 & 338 & 25.6 & 0.591 & 0.76 & - & - \\
        {\model }  & 6 & 8 & 96 & 8.7 & 306 & 22.4 & 0.634 & 1.16 & - & - \\
        {\model }  & 6 & 12 & 96 & 6.8 & 294 & 21.3  & 0.653 & 1.45 & \bf{376} & \bf{10.9} \\
        {\model }  & 6 & 16 & 64 & 4.2 & \bf{283} & \bf{20.5} & \bf{0.662} & \bf{1.67} & \bf{371} & \bf{11.2} \\
     
      \bottomrule
    \end{tabular}
    \end{adjustbox}
    \caption{Results on the \ocd~2M training dataset and ablation studies for \model{}~model variations. The validation results are averaged across the four OC20 Validation set splits. All \model{}~models are trained on 16 GPUs for 12 epochs with the learning rate reduced by 0.3 at 7, 9, and 11 epochs. Batch sizes vary based on the number of instances that can be fit in 32GB RAM.}
    \label{tab:comp-ablation}
\end{table*}

Following recent approaches \cite{gasteiger2021gemnet,shuaibi2021rotation,zitnick2022spherical}, we evaluate our model on the large scale Open Catalyst 2020 (OC20) and Open Catalyst 2022 (OC22) datasets \citep{OC20} containing 130M and 8M training examples respectively. Most training examples are obtained from relaxation trajectories of catalysts with adsorbate molecules. That is, a molecule is placed near the catalyst's surface, and the atom positions are relaxed until a local energy minimum is found. Some OC22 examples do not contain an adsorbate. The datasets were specifically designed to aid in the discovery of new catalysts for helping address climate change \cite{zitnick2020introduction} and is released under a Creative Commons Attribution 4.0 License. We evaluate our model on all test tasks, and report ablation studies on the smaller OC20 2M dataset.

\subsection{Implementation details}

Unless otherwise stated, all models are trained with 12 layers, $C=128$ channels, $H=256$ hidden units, $L=6$ degrees, and $M=2$ orders. Neighbors were determined by selecting the 20 closest atoms with a distance less than $12\text{\AA}$. The AdamW optimizer with a fixed learning rate schedule was used for all training runs. For the OC20 2M training dataset, the learning rate was 0.0008 and dropped by 0.3 at 7, 9, 11 epochs. Training was stopped at 12 epochs. For the All and All+MD runs, the same initial learning rate was used with drops after 41M, 52M and 62M training examples. Training was stopped before a complete epoch was completed. The force loss had a coefficient of 100, and the energy loss a coefficient of 2 (2M) or 4 (All, All+MD). All training runs used data parallelism and PyTorch’s Automatic Mixed Precision (AMP). The model source code and checkpoints are publicly available in the Open Catalyst Github repo under an MIT
license.

\subsection{Results}

We compare our \model~model against state-of-the-art approaches in Table \ref{tab:comp-all} on the OC20 All and OC20 All+MD datasets and in Table \ref{tab:comp-oc22} on the OC22 dataset. The All dataset contains 130M training examples, the MD dataset contains 38M examples and OC22 contains 8M examples. Results are shown for the structure to energy and forces (S2EF), initial structure to relaxed structure (IS2RS) and initial structure to relaxed energy (IS2RE) tasks. For IS2RS and IS2RE tasks we use the relaxation approach which uses the model to estimate the atom forces to iteratively update the atom positions until a local energy minimum is found (the forces are close to zero). Relaxations are performed for 200 time steps using an LBFGS implementation from the Open Catalyst repository. 

The results show \model~outperforming other models in tasks that require high-fidelity directional information that the higher degrees $L$ provide, such as force MAE and force cosine. The force MAE improves by $9\%$ for OC20 All+MD and $21\%$ for OC22 ID. Notably, the AFbT metric for IS2RS on OC20 measures how often relaxed structures are found, as verified by DFT, using the ML model's force estimates. This is a critical capability for ML models used to replace DFT \cite{lan2022adsorbml}. On this metric, \model~achieves over a $15\%$ increase in accuracy. Energy prediction, which benefits from longer range reasoning between nodes, is on par with SCN and GemNet-OC on OC20 MD+All and OC22. 

\subsection{Ablation studies}
In Section \ref{sec:eq_eff}, we showed our novel approach drastically reduces the cost of computing the tensor product. Here, we validate this claim by implementing two mathematically identical equivariat GNNs; one using \textit{e3nn} PyTorch library \citep{geiger2022e3nn} and the other one our efficient implementation.  In Figure \ref{fig:e3nn_comp} we compare the GPU memory allocated at training time and the Time / Epoch over the maximal degree of the irreps. We conclude that our method can reduce the computational cost by an order of magnitude as we scale $L$.

Additionally, we perform ablation studies across three variants of our model; varying the degree $L$, the order $M$, and the number of layers $K$ in Table \ref{tab:comp-ablation}. We see that energy MAE is more sensitive to the number of layers $K$, while force MAE is more sensitive to the degree $L$. We hypothesize this is due to the energy being more dependent on the entire structure's configuration so more layers are helpful, while forces are more directionally dependent so high degrees of $L$ (higher directional resolution) are beneficial. Increasing $M$ improves results up to $M=2$. Higher values may not increase accuracy since $M$ controls the resolution of the node representations perpendicular to the edges' directions, which may be less useful to the network. Finally, we show results when the non-linear point-wise activation function, Equation (\ref{eqn:pointwise}), is removed from message passing (\model~linear). We see accuracies decrease, but are still comparable to models with lower degrees or fewer layers.

The run times of the different model variations are also shown. Faster runtimes than those shown are possible for smaller models, if the batch sizes had been increased. Overall runtimes are similar to those of the SCN model.

\section{Discussion}

Research in atomic modeling is critical for addressing many of the world's problems, especially with respect to our challenges with climate change. However, advances in chemistry can have both positive and negative impact as has been demonstrated in other chemistry breakthroughs. One cautionary example is the Haber-Bosch process \cite{hager2009alchemy} for ammonia production that enabled the world to feed itself, but has led to over fertilization and ocean dead zones. We hope to inspire positive uses by demonstrating our results on datasets such as OC20 \cite{OC20}.

The constraints imposed by the $SO(3)$-equivariance not only limit the use of linear functions but also non-linearities. We utilize the point-wise spherical non-linearity since it is very expressive, as it processes the full spectrum of the spherical harmonics. Most other equivariant non-linear functions are applied to the norm of irreps and scalar coefficients. For point-wise spherical non-linearities the discrete transformation of the irreps to the sphere makes the operation quasi-equivariant and can be computationally expensive if equivariance to numerical precision is desired, Section \ref{app:quasi}. 

In conclusion, we unveil the relationship between $SO(3)$ and $SO(2)$ convolutions and how this can led to dramatic improvements in computational efficiency. We demonstrate the approach using our \model~model that achieves state-of-the-art performance on atomic modeling tasks such as force prediction. 

\section*{Acknowledgements}
The authors thank Abhishek Das, Janice Lan, Brandon Wood, Mark Tygert and Gabriele Corso for valuable feedback and insightful discussions.

\bibliography{eSCN}
\bibliographystyle{icml2023}

%%%%%%%%%%%%%%%%%%%%%%%%%%%%%%%%%%%%%%%%%%%%%%%%%%%%%%%%%%%%%%%%%%%%%%%%%%%%%%%
%%%%%%%%%%%%%%%%%%%%%%%%%%%%%%%%%%%%%%%%%%%%%%%%%%%%%%%%%%%%%%%%%%%%%%%%%%%%%%%
% APPENDIX
%%%%%%%%%%%%%%%%%%%%%%%%%%%%%%%%%%%%%%%%%%%%%%%%%%%%%%%%%%%%%%%%%%%%%%%%%%%%%%%
%%%%%%%%%%%%%%%%%%%%%%%%%%%%%%%%%%%%%%%%%%%%%%%%%%%%%%%%%%%%%%%%%%%%%%%%%%%%%%%

\newpage
\appendix
\onecolumn
\section*{Appendix: Table of contents}
\begin{itemize}
    \item[] \ref{app:proofs} Proofs
    \item[] \ref{app:maths} Mathematical foundations of $SO(3)$ and $SO(2)$ irreps
    \item[] \ref{app:comp} Computational comparison with e3nn tensor product
    \item[] \ref{app:quasi} Analysis on the quasi-equivariance of the spherical activation function
    \item[] \ref{app:sample} Sample efficiency of \model
    
\end{itemize}

\section{Proofs} \label{app:proofs}
\subsection{Proof of Proposition \ref{prop:C-G}}\label{proof:C-G}
\begin{figure*}[htb]
\begin{center}
    \includegraphics[width=0.55\textwidth]{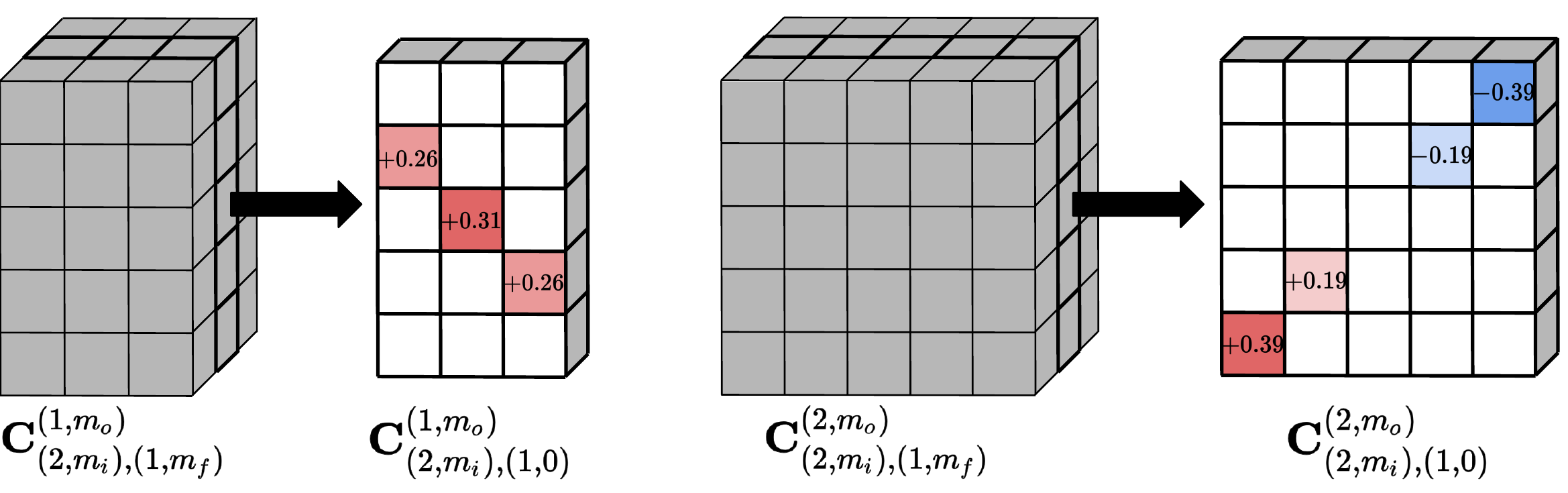}
\vspace{-6pt}
\caption{Visual representation of the Clebsch-Gordan matrices $\mathbf{C}^{(1, m_o)}_{(2, m_i), (1, m_f)} \in \mathbb{R}^{5 \times 3 \times 3}$ and $\mathbf{C}^{(2, m_o)}_{(2, m_i), (1, m_f)} \in \mathbb{R}^{5 \times 3 \times 5}$. 
  The computational cost of the tensor product is reduced by aligning the direction $\hat{r}$ to the y-axis as we retain only the coefficients with $m_f = 0$ during the multiplication. We illustrate the sparsity of the Clebsch-Gordan matrices as stated in Proposition \ref{prop:C-G}.}
\end{center}
\end{figure*}

We start the proof noting that the Clebsch-Gordan coefficients used in the paper differ from the standard coefficients used in physics. The Clebsch-Gordan matrix used in quantum mechanics is the change-of-basis which decomposes the $SU(2)$ tensor product between irreducible representations.
However, the $SO(3)$ irreps are related to the $SU(2)$ irreps as follows:
\begin{equation}\label{eqn:C-G_complex}
    \mathbf{x}_{m}^{(l)} = \begin{cases}
     \frac{i}{\sqrt{2}}\left( \mathbf{z}_{m}^{(l)} - (-1)^m \mathbf{z}_{-m}^{(l)}\right) & \text{if } m < 0 
    \\
     \mathbf{z}_{0}^{(l)} & \text{if } m = 0 
    \\
     \frac{1}{\sqrt{2}}\left( \mathbf{z}_{m}^{(l)} + (-1)^m \mathbf{z}_{-m}^{(l)}\right) & \text{if } m > 0 
\end{cases}
\end{equation}
where $\mathbf{z}$ is an $SU(2)$ irrep of degree $l$ and $\mathbf{x}$ an $SO(3)$ irrep of degree $l$.

Additionally, we denote the $SU(2)$ Clebsch-Gordan coefficients by $\mathbf{C}^{* (l_o, m_o)}_{(l_i, m_i), (l_f, m_f)}$ .
A well-known property \citep{weisstein2003clebsch} of $\mathbf{C}^{* (l_o, m_o)}_{(l_i, m_i), (l_f, m_f)}$ is that they are non-zero only if $m_o = m_i + m_f$ and the following holds:
\begin{equation}
    \mathbf{C}^{* (l_o, m_o)}_{(l_i, m_i), (l_f, m_f)} = (-1)^{l_i+l_f+l_o} \mathbf{C}^{*(l_o, -m_o)}_{(l_i, -m_i), (l_f, -m_f)}
\end{equation}
Therefore by substituting $m_f = 0$ we obtain that $\mathbf{C}^{* (l_o, m_o)}_{(l_i, m_i), (l_f, 0)}$ is non-zero only if $m_o = m_i$ and the the following holds for any $m$:
\begin{equation}
    \mathbf{C}^{* (l_o, m)}_{(l_i, m), (l_f, 0)} = (-1)^{l_i+l_f+l_o} \mathbf{C}^{* (l_o, -m)}_{(l_i, -m), (l_f, 0)}
\end{equation}
Now, we can relate the $SU(2)$ Clebsh-Gordan coefficients to the $SO(3)$ ones using a change-of-basis directly derived from Equation (\ref{eqn:C-G_complex}).
We fix $m_o=0$ and note that the components $\{-m, m\}$ of an $SU(2)$ irrep get rearranged to the components $\{-m, m\}$ of an $SO(3)$ irrep as follows; for $m \ne 0$:
\begin{equation}
\begin{split}
    \begin{pmatrix}
\mathbf{C}^{ (l_o, -m)}_{(l_i, -m), (l_f, 0)} & \mathbf{C}^{ (l_o, m)}_{(l_i, -m), (l_f, 0)} \\
\mathbf{C}^{ (l_o, -m)}_{(l_i, m), (l_f, 0)} & \mathbf{C}^{ (l_o, m)}_{(l_i, m), (l_f, 0)}
\end{pmatrix} = \begin{pmatrix}
\frac{i}{\sqrt{2}} & \frac{-i (-1)^m}{\sqrt{2}} \\
\frac{(-1)^m}{\sqrt{2}} & \frac{1}{\sqrt{2}}
\end{pmatrix} \begin{pmatrix}
\mathbf{C}^{* (l_o, -m)}_{(l_i, -m), (l_f, 0)} & \mathbf{C}^{* (l_o, m)}_{(l_i, -m), (l_f, 0)} \\
\mathbf{C}^{* (l_o, -m)}_{(l_i, m), (l_f, 0)} & \mathbf{C}^{* (l_o, m)}_{(l_i, m), (l_f, 0)}
\end{pmatrix} \begin{pmatrix}
\frac{i}{\sqrt{2}} & \frac{-i (-1)^m}{\sqrt{2}} \\
\frac{(-1)^m}{\sqrt{2}} & \frac{1}{\sqrt{2}}
\end{pmatrix}^{-1} = 
\\
= \begin{cases}
    \begin{pmatrix}
\mathbf{C}^{* (l_o, m)}_{(l_i, m), (l_f, 0)} & 0 \\
0 & \mathbf{C}^{* (l_o, m)}_{(l_i, m), (l_f, 0)}
\end{pmatrix} & \text{  if }  l_i +l_f + l_o \text{ is even} 
     \\
\begin{pmatrix}
0 & 2 (-1)^m \mathbf{C}^{* (l_o, m)}_{(l_i, -m), (l_f, 0)} \\
-2 (-1)^m \mathbf{C}^{* (l_o, m)}_{(l_i, -m), (l_f, 0)} & 0 \end{pmatrix}  & \text{  if } l_i +l_f + l_o \text{ is odd} 
\end{cases}
\end{split}
\end{equation}
for $m=0$:
\begin{equation}
     \mathbf{C}^{ (l_o, 0)}_{(l_i, 0), (l_f, 0)} = \mathbf{C}^{* (l_o, 0)}_{(l_i, 0), (l_f, 0)}
\end{equation}
Moreover $\mathbf{C}^{* (l_o, 0)}_{(l_i, 0), (l_f, 0)} = (-1)^{l_i + l_f + l_o} \mathbf{C}^{* (l_o, 0)}_{(l_i, 0), (l_f, 0)}$, so $\mathbf{C}^{ (l_o, 0)}_{(l_i, 0), (l_f, 0)} = 0$ for $l_i + l_f + l_o$ odd.

Additionally, we note that $\mathbf{C}^{* (l_o, m_o)}_{(l_i, m_i), (l_f, m_f)} =0$ if $|m_o|\ne |m_i|$ implies that the same is true for  $\mathbf{C}^{(l_o, m_o)}_{(l_i, m_i), (l_f, m_f)}$.

We finally conclude that $\mathbf{C}^{(l_o, m_o)}_{(l_i, m_i), (l_f, 0)}$ is non-zero only for $m_o = \pm m_i$, and  $\mathbf{C}^{(l_o, m)}_{(l_i, m), (l_f, 0)} = \mathbf{C}^{(l_o, -m)}_{(l_i, -m), (l_f, 0)}$ and $\mathbf{C}^{(l_o, m)}_{(l_i, m), (l_f, 0)} = - \mathbf{C}^{(l_o, -m)}_{(l_i, -m), (l_f, 0)}$  for any $m$. Moreover, for any $m \ne 0$, $\mathbf{C}^{(l_o, m)}_{(l_i, m), (l_f, 0)} = 0$ when $l_i + l_f + l_o$ odd and $\mathbf{C}^{(l_o, m)}_{(l_i, -m), (l_f, 0)} = 0$ when $l_i + l_f + l_o$ even; for $m=0$, $\mathbf{C}^{(l_o, 0)}_{(l_i, 0), (l_f, 0)} = 0$ when $l_i + l_f + l_o$ odd.

This concludes the proof of the Proposition.

\subsection{Proof that there exists a linear bijection between $(\mathbf{h}_{l_i, l_f, l_o})_{|l_i - l_o| \le l_f \le l_i + l_o}$ and $(\tilde{\mathbf{h}}_{m}^{(l_i, l_o)})_{-min(l_i, l_o) \le m \le min(l_i, l_o)}$}\label{app:bij}
\begin{figure*}[htb]
\begin{center}
    \includegraphics[width=0.65\textwidth]{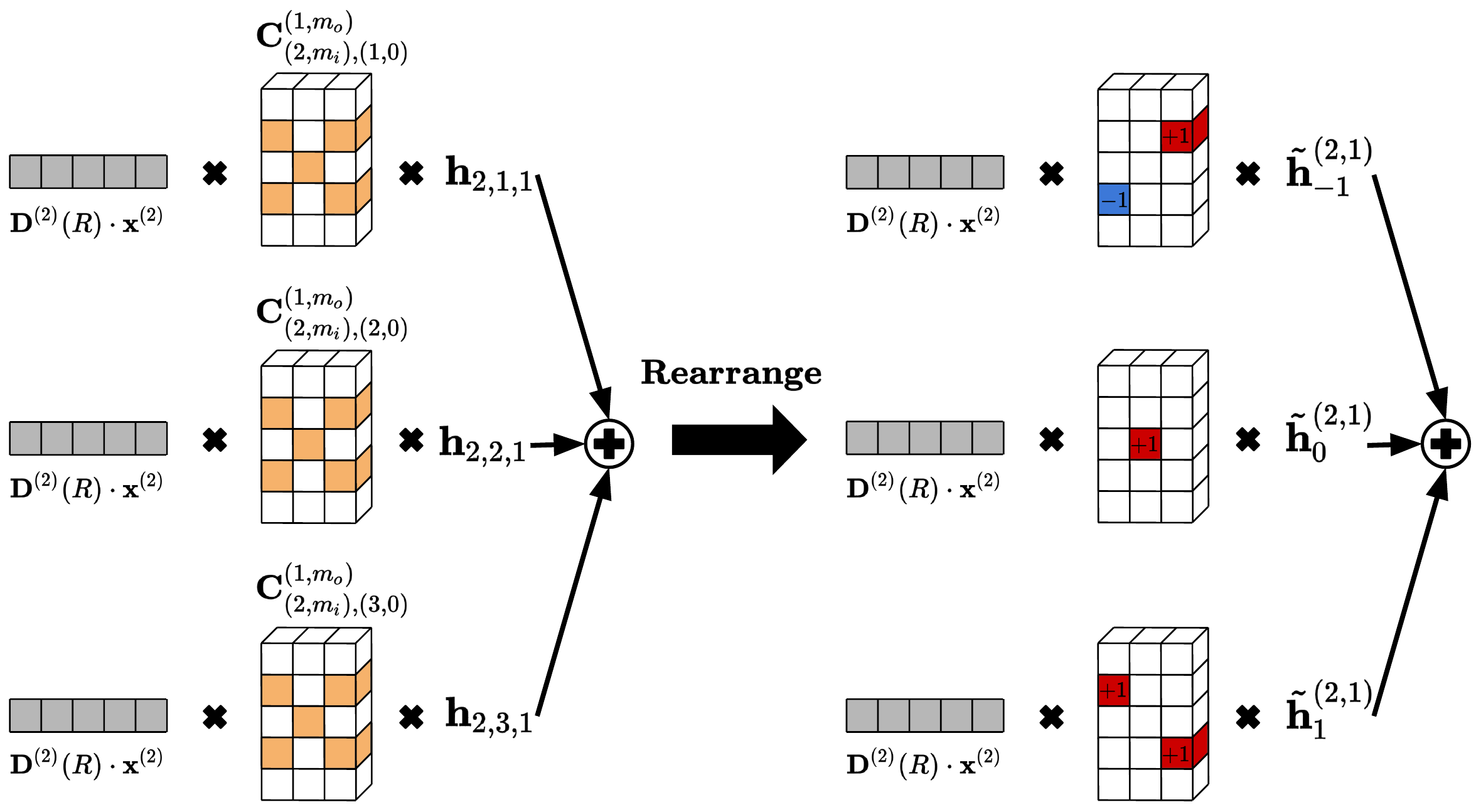}
\vspace{-6pt}
\caption{Visual representation of the reparametrization of the tensor products $\otimes^{1}_{2, 0}, \otimes^{1}_{2, 1}, \otimes^{1}_{2, 2}$ between $\mathbf{D}^{(2)}(R)\mathbf{x}^{(2)} \in \mathbb{R}^5$ and the directions $\mathbf{Y}^{(0)}(0, 1, 0), \mathbf{Y}^{(1)}(0, 1, 0), \mathbf{Y}^{(2)}(0, 1, 0)$. The coefficients $\mathbf{h}_{2, 1, 1}, \mathbf{h}_{2, 2, 1}, \mathbf{h}_{2, 3, 1}$ get rearranged to $\tilde{\mathbf{h}}_{-1}^{(2, 1)}, \tilde{\mathbf{h}}_{0}^{(2, 1)}, \tilde{\mathbf{h}}_{1}^{(2, 1)}$.}
\end{center}
\end{figure*}

We start by noting that the $SO(3)$ Clebsch-Gordan coefficients are related to the $SU(2)$ ones with a linear invertible transformation. Therefore without loss of generality we can assume that $\mathbf{C}^{(l_o, m_o)}_{(l_i, m_i), (l_f, m_f)}$ are the ones used in quantum mechanics.

By definition, $\tilde{\mathbf{h}}^{(l_i, l_o)}_{m} = \sum_{l_f}  \left( \mathbf{h}_{l_i, l_f, l_o}  (\mathbf{c}_{l_i,l_f,l_o})_m \right)$, which specifies a linear surjection from $(\mathbf{h}_{l_i, l_f, l_o})_{l_f}$ to $(\tilde{\mathbf{h}}^{(l_i, l_o)}_{m})_m$ for a fixed value of $l_i$ and $l_o$.

We now prove that the aforementioned map from $\mathbf{h}_{l_i, l_f, l_o}$ to $\tilde{\mathbf{h}}^{(l_i, l_o)}_{m}$ is injective. Let's suppose exists a set of $(\mathbf{h}_{l_i, l_f, l_o})_{l_f}$ such that  $\tilde{\mathbf{h}}^{(l_i, l_o)}_{m} = 0$ for any $l_i, l_o, m$. From Equation (\ref{eqn:SO2_conv}) we conclude that:
\begin{equation*}
    \sum_{l_f} \tilde{\mathbf{x}}_s^{(l_i)} \otimes^{l_o}_{l_i, l_f} \mathbf{h}_{l_i, l_f, l_o}  \mathbf{\delta}^{(l_f)} = 0 \quad \implies \quad \sum_{l_f} \left(\mathbf{h}_{l_i, l_f, l_o} \mathbf{C}^{(l_o, :)}_{(l_i, :), (l_f, 0)} \right) \cdot \tilde{\mathbf{x}}_s^{(l_i)} = 0
\end{equation*}
where $\sum_{l_f} \left(\mathbf{h}_{l_i, l_f, l_o} \mathbf{C}^{(l_o, :)}_{(l_i, :), (l_f, 0)} \right)$ is a 2D matrix which multiplies the vector $\tilde{\mathbf{x}}_s^{(l_i)}$. The above expression is $0$ for any value of $\tilde{\mathbf{x}}_s^{(l_i)}$ which implies that the matrix $\sum_{l_f} \left(\mathbf{h}_{l_i, l_f, l_o} \mathbf{C}^{(l_o, :)}_{(l_i, :), (l_f, 0)} \right)$ is zero, therefore for any value of $m_i$ and $m_o$:
\begin{equation}\label{eqn:proof_C-G}
    \sum_{l_f} \left(\mathbf{h}_{l_i, l_f, l_o} \mathbf{C}^{(l_o, m_o)}_{(l_i, m_i), (l_f, 0)} \right) = 0
\end{equation}
Additionally, the Clebsch-Gordan coefficients satisfy the following properties \citep{weisstein2003clebsch}; the first one is an orthogonality identity, the second one a permutation identity:
\begin{equation}\label{eqn:id1}
    \sum_{m_i, m_f} \left( \mathbf{C}^{(l_o, m_o)}_{(l_i, m_i),(l_f, m_f)} \mathbf{C}^{(l_o', m_o')}_{(l_i, m_i),(l_f, m_f)} \right) = \delta_{m_o m_o'} \delta_{l_o l_o'}
\end{equation}
\begin{equation}\label{eqn:id2}
    \mathbf{C}^{(l_o, m_o)}_{(l_i, m_i),(l_f, m_f)} = \mathbf{C}^{(l_f, -m_f)}_{(l_i, m_i),(l_o, -m_o)} (-1)^{l_i + m_i} \sqrt{\frac{2l_o + 1}{2l_f + 1}}
\end{equation}
By taking the sum of Equation (\ref{eqn:proof_C-G}) over $m_i, m_o$ for a fixed value of $l_f'$ we obtain:
\begin{equation}
\begin{split}
&0 = \sum_{m_i, m_o} \mathbf{C}^{(l_o, m_o)}_{(l_i, m_i), (l_f', 0)} \sum_{l_f} \left(\mathbf{h}_{l_i, l_f, l_o} \mathbf{C}^{(l_o, m_o)}_{(l_i, m_i), (l_f, 0)} \right) = \sum_{l_f} \mathbf{h}_{l_i, l_f, l_o} \sum_{m_i, m_o} \left(\mathbf{C}^{(l_o, m_o)}_{(l_i, m_i), (l_f', 0)}  \mathbf{C}^{(l_o, m_o)}_{(l_i, m_i), (l_f, 0)} \right) = 
\\
&\sum_{l_f} \mathbf{h}_{l_i, l_f, l_o} \sum_{m_i, m_o} \left(\mathbf{C}^{(l_o, m_o)}_{(l_i, m_i), (l_f', 0)}  \mathbf{C}^{(l_o, m_o)}_{(l_i, m_i), (l_f, 0)} \right) = \sum_{l_f} \mathbf{h}_{l_i, l_f, l_o} \sum_{m_i, m_o} \sqrt{\frac{2l_o + 1}{2l_f + 1}}\left(\mathbf{C}^{(l_f', 0)}_{(l_i, m_i), (l_o, -m_o)}  \mathbf{C}^{(l_f, 0)}_{(l_i, m_f), (l_o, -m_o)} \right) = 
\\
&\sum_{l_f} \mathbf{h}_{l_i, l_f, l_o} \sqrt{\frac{2l_o + 1}{2l_f + 1}}  \delta_{l_f l_f'} = \mathbf{h}_{l_i, l_f', l_o} \sqrt{\frac{2l_o + 1}{2l_f' + 1}}
\end{split}
\end{equation}
where we substitute Equations (\ref{eqn:id1}) and (\ref{eqn:id1}) in the 5th and 4th equality.
Therefore we conclude that $\mathbf{h}_{l_i, l_f', l_o} = 0$ for any value of $l_f'$, which implies that the linear map is injective.

We conclude the proof noting that a linear map between finite dimensional vector spaces which is both surjective and injective is bijective.
\subsection{Proof of Theorem \ref{th:so2}}
We first prove that there exists a projection map satisfying $\mathbf{w}_0 = (\mathbf{x})_0$ and $\mathbf{w}_k = (\mathbf{x})_{(-k, k)}$ for each $k \ne 0$.
We note that, if we define $\mathbf{w}$ in terms of $\mathbf{x}$ as above, and we let $\mathbf{W}_0, ... , \mathbf{W}_l$ to be the spaces $\langle (\mathbf{x})_0 \rangle, \langle (\mathbf{x})_1, (\mathbf{x})_{-1} \rangle, \langle (\mathbf{x})_2, (\mathbf{x})_{-2} \rangle, ... ,  \langle (\mathbf{x})_l, (\mathbf{x})_{-l} \rangle$ then $\bigoplus_k \mathbf{W}_k = \mathbf{V}_l$. We are left to prove that the representations $\tau_i : SO(2) \rightarrow GL(\mathbf{W}_i)$ are irreducible. A sufficient condition to prove it is to show that $\mathbf{w}_k$ transforms like a circular harmonic of degree $k$ under the action $\rho_{|G_{\hat{n}}}(g)$. We note that this has already been proved in the main paper, which concludes the proof.

Lastly, we note that the projection map is invertible since, by definition, the linear change-of-basis map relating $\mathbf{w}_0, ... , \mathbf{w}_l$ and $\mathbf{x}$ is invertible.

\section{Mathematical foundations of $SO(3)$ and $SO(2)$ irreps}\label{app:maths}
\subsection{General introduction to equivariance}

Group theory is a formalism to axiomatically define symmetries. A group $G$ is a set of elements with an operation $\bullet_G : G \times G \rightarrow G$ which satisfies the following properties:
\begin{enumerate}
    \item There exists $e \in G$ such that $\forall g \in G$ $g\bullet_G e = e \bullet_G g = g$
    \item For all $g_1, g_2, g_3 \in G$ we have $g_1 \bullet_G (g_2 \bullet_G g_3) = (g_1 \bullet_G g_2) \bullet_G g_3$
    \item For all $g \in G$ there exists $g^{-1} \in G$ such that $g\bullet_Gg^{-1} = g^{-1} \bullet_G g = e$
\end{enumerate}
We can define infinitely many groups but the ones we are interested in are $\mathbb{T}$, the translation group, and most importantly $SO(3)$, the rotation group of $\mathbb{R}^3$.
The elements of $SO(3)$ should be regarded as rotations in $\mathbb{R}^3$ and the group operation $\bullet_{SO(3)}$ as the composition of $2$ rotations.
In our context the symmetries transform a surrounding space: for $SO(3)$ this space is $\mathbb{R}^3$. Group action is a way to axiomatically define this operation.
A group action on a vector space $V$ is an operation $* : G \times V \rightarrow V$ which satisfies:
\begin{enumerate}
    \item Exists $e \in G$ such that $\forall \mathbf{v} \in V$   $e * \mathbf{v} = \mathbf{v}$
    \item $\forall g_1, g_2 \in G$ we have $(g_1 \bullet_G g_2) * \mathbf{v} = (g_1 * (g_2 * \mathbf{v}))$
    \item $\forall \mathbf{v}, \mathbf{w} \in V, \lambda \in \mathbb{R}, g \in G$ we have $g * (\lambda \mathbf{v} + \mathbf{w}) = \lambda g * \mathbf{v} + g * \mathbf{w}$
\end{enumerate}
The canonical action of $SO(3)$ takes the symmetry (i.e. the rotation), an element of $\mathbb{R}^3$ and rotates it accordingly.

Given two vector spaces $V, W$ (i.e. $\mathbb{R}^n$, $\mathbb{R}^m$ for some $n$ and $m$) and a group $G$, we say that $F: V \rightarrow W$ is:
\begin{enumerate}
    \item $G$-\textbf{Invariant} if $\forall \mathbf{v} \in V $    $g \in G \quad$  $F(g * \mathbf{v}) = F(v)$
    \item $G$-\textbf{Equivariant} if $\forall \mathbf{v} \in V $  $g \in G\quad$  $F(g * \mathbf{v}) = g * F(\mathbf{v})$
\end{enumerate}
In the OC-20 task the energy is invariant for both $\mathbb{T}$ and $SO(3)$, per-atom forces are invariant to $\mathbb{T}$ and equivariant to $SO(3)$.

\subsection{Irreducible representations of SO(3) and SO(2)}

We limit our study of symmetries to $SO(3)$ and $SO(2)$, the group of 3D and 2D rotations, which are canonically acting on $\mathbb{R}^3$ and $\mathbb{R}^2$. In this section, we classify how to extend the $SO(3)$ and $SO(2)$ action to other vector spaces $V = \mathbb{R}^n$ for some integer $n$.

We define the concept of group homomorphism to rephrase the the problem from another point of view. Given two groups $G, H$ we say $\rho: G \rightarrow H$ is a group homomorphism if it satisfies $\forall g_1, g_2 \in G$ $\rho(g_1 \bullet_{G} g_2) = \rho(g_1) \bullet_{H} \rho(g_2)$.

We note that an action of a group $G$ on a vector space $V$ is mathematically equivalent to the existence of a group homomorphism $\rho : G \rightarrow GL(V)$ where $GL(V)$ is the general linear group, the group of invertible matrices on $V$. We define $\rho$ to be a representation of $SO(3)$ which acts as matrix multiplication on the vector space $V = \mathbb{R}^n$ as follows:
\begin{equation}
    g * \mathbf{v} = \rho(g)\mathbf{v}  \quad \forall g \in SO(3)
\end{equation}
Additionally, we define the concept of irreducible representations. Specifically, a representaion $\rho : SO(3) \rightarrow GL(V)$ is reducible if there exists a non-trivial subspace $W \le V$ such that $\forall g \in SO(3)$, $\mathbf{w} \in W$ we have that $\rho(g)\mathbf{w} \in W$; otherwise we say the representation is irreducible.

We finally state the classification of the $SO(3)$ and $SO(2)$ \textit{irreducible} representations. 

For $SO(3)$, only for $n = 2l + 1$ odd there exists an irreducible representation $\rho : SO(3) \rightarrow V = \mathbb{R}^n$. Moreover $\rho$ is unique up to a change of basis; we decide to remove the degeneracy arising from the possible change-of-basis by aligning the rotations' generator to the \textit{y} axis (as in the e3nn library). We define the Wigner-$D$ matrix of degree $l$ to be $\mathbf{D}^{(l)} : SO(3) \rightarrow \mathbb{R}^{2l + 1}$:
\begin{equation}\label{eq:D_mat}
    \mathbf{D}^{(l)}(g) : = \rho(g) \quad  \forall g \in SO(3)
\end{equation}
Therefore we let $\mathbf{V}_{l} = \mathbb{R}^{2l + 1}$ to be a $2l+1$ dimensional vector space equipped with the $SO(3)$ action: the multiplication by a Wigner-D matrix. The elements of $\mathbf{V}_{l}$ are called $SO(3)$-irreps of degree $l$.

Analogously, there exists a unique $\rho : SO(2) \rightarrow W = \mathbb{R}^n$ irreducible representation for $n=1$, infinitely many for $n=2$ and none for $n > 2$.  We enumerate the possible representations with an index $k \in \mathbb{R}_{+}$ where $W = \mathbb{R}$ for $k = 0$ and $W = \mathbb{R}^2$ for $k > 0$ and we define $\mathbf{C}^{(k)} : SO(2) \rightarrow W$ to be the irreducible representation indexed by $k$. It satisfies
\begin{equation}\label{eq:C_mat}
    \mathbf{C}^{(k)}(g) : = \rho(g) = g^k \quad  \forall g \in SO(2)
\end{equation}
We define $\mathbf{W}_{k} = \begin{cases}
     \mathbb{R} & \text{if } k = 0 
    \\
     \mathbb{R}^2  & \text{if } k > 0
\end{cases} \quad$ to be the vector space of the $SO(2)$-irreps equipped with the $SO(2)$ action: the multiplication by the matrix of Equation \ref{eq:C_mat}.

\subsection{Spherical and Circular Harmonics}

The irreps have been intuitively defined starting from the first principles of symmetries. Nonetheless it can be hard to have intuition on the resulting space since we have added a few layers of abstraction. In this Section we introduce a way to build geometrical intuition by linking them to the spherical and circular harmonics.

The real spherical harmonic of degree $l$ is a function $\mathbf{Y}^{(l)} : S^2 \rightarrow \mathbf{V}_l$. The set of components of $\mathbf{Y}^{(l)}$ produces a basis for the homogeneous polynomial of degree $l$ on the sphere.

Analogously, the real circular harmonic of degree $k$ is a function $\mathbf{B}^{(k)} : S \rightarrow \mathbf{W}_k$, whose components produce a basis for the homogeneous polynomial of degree $k$ on the circle.

Most importantly they satisfy the following expressions:
\begin{equation}
\begin{split}
    \mathbf{Y}^{(l)}(g\hat{\mathbf{n}}) = \mathbf{D}^{(l)}(g) \cdot \mathbf{Y}^{(l)}(\hat{\mathbf{n}}) \quad   \hat{\mathbf{n}} \in S^2 \quad g \in SO(3)
     \\
    \mathbf{B}^{(k)}(g\hat{\mathbf{n}}) = \mathbf{C}^{(k)}(g) \mathbf{B}^{(k)}(\hat{\mathbf{n}}) \quad   \hat{\mathbf{n}} \in S \quad g \in SO(2)
\end{split}
\end{equation}
We can make sense of a set of spherical harmonics $\oplus_i \mathbf{x}_i \in \bigoplus_i \mathbf{V}_i$ by defining a function on the sphere $F_{\mathbf{x}}: S^2 \rightarrow \mathbb{R}$ such that $\forall \hat{\mathbf{n}} \in S^2 \quad F_{\mathbf{x}}(\hat{\mathbf{n}}) := \sum_l \mathbf{x}_l \cdot \mathbf{Y}^{(l)}(\hat{\mathbf{n}})$.
In this setting, acting on $\oplus_i \mathbf{x}_i$ with a rotation matrix $R \in SO(3)$ corresponds to rotating the frame of reference of the sphere. This shows that the $SO(3)$-irreps can be treated as coefficients of the spherical harmonics.
Analogously this is true for the $SO(2)$-irreps and the circular harmonics on a circle.

\subsection{Projection to SO(2)}\label{sec:proj}

Given an irreducible representation $\rho : SO(3) \rightarrow GL(\mathbf{V}_{l})$ and a direction $\hat{\mathbf{n}} \in \mathbb{R}^3$ we define a projection as follows.
\begin{definition}\label{def:so2}
    {
    The projection of $\rho$ along $\hat{\mathbf{n}}$ is $\rho_{\hat{\mathbf{n}}} : G_{\hat{\mathbf{n}}} \rightarrow GL(\mathbf{V}_{l})$ where $G_{\hat{\mathbf{n}}}$ is the subgroup of rotations around $\hat{\mathbf{n}}$ and $\rho_{\hat{\mathbf{n}}} = \rho_{|G_{\hat{\mathbf{n}}}}$ is obtained by restricting the domain of the representation.
    }
\end{definition}
We reckon that $\rho_{\hat{\mathbf{n}}}$ is a representation of $SO(2)$ since $G_{\hat{\mathbf{n}}}$ is isomorphic to $SO(2)$ but it is not necessarily irreducible. Nonetheless, we can apply a change of basis and reduce it to
\begin{equation}
\begin{split}
    \tau_0 : SO(2) \rightarrow GL(\mathbf{W}_0)
     \\...\\
    \tau_l : SO(2) \rightarrow GL(\mathbf{W}_l)
\end{split}
\end{equation}
such that $\bigoplus_i \mathbf{W}_{i} = \mathbf{V}_{l}$ and $\tau_i$ is an irreducible representation of $SO(2)$ for any $i$.
This induces a projection map 
\begin{equation}
\mathlarger{\Pi}_{\hat{\mathbf{n}}}: \mathbf{V}_l \rightarrow \bigoplus_{i=0}^l  \mathbf{W}_i
\end{equation}
We note that the projection map is unique up to the change of basis of $\mathbf{W}_i$.
We provide in Theorem \ref{th:so2} an explicit way to compute a projection map for $\hat{\mathbf{n}} = (0, 1, 0)$ and we prove that such a map is invertible.
\begin{theorem}\label{th:so2}
    {Let $\mathbf{x} \in \mathbf{V}_l$ and $\hat{\mathbf{n}} = (0, 1, 0)$ then there exists a projection map $\mathlarger{\Pi}_{\hat{\mathbf{n}}}(\mathbf{x}) = \oplus_{i=0}^l \mathbf{w}_i$ such that $\mathbf{w}_i \in \mathbf{W}_i$, $\mathbf{w}_0 = (\mathbf{x})_0 $ and $\mathbf{w}_k = (\mathbf{x})_{(-k, k)}$ for each $1 \le k \le l$

    Moreover if $\oplus_i \mathbf{w}_i \in \bigoplus_{i=0}^l  \mathbf{W}_i$ then  $\mathlarger{\Pi}_{\hat{\mathbf{n}}}^{-1}(\oplus_i \mathbf{w}_i) = \mathbf{x} \in \mathbf{V}_l$ so that: $(\mathbf{x})_0 = \mathbf{w}_0$ and $\mathbf{x}_{k} = (\mathbf{w}_{|k|})_{sign(k)} $ for each $1 \le k \le l$
    }
\end{theorem}

\section{Computational comparison with e3nn tensor product}\label{app:comp}

We study the computational advantage of our method from a theoretical point of view and we empirically validate our study.

We fix $L$ and study the computational complexity of a network with $C$ channels for each $0 \le l \le L$. The naive implementation calculates a tensor product $\otimes^{l_o}_{l_i, l_f}$ for $0 \le l_i \le L$, $0 \le l_o \le L$ and $|l_i -l_o| \le l_f \le l_i + l_o$ followed by a linear self-interaction expanding the number of channels of each degree $l_o$ back to $C$. The number of tensor product operations scale with $O(C \cdot L^3)$, each of which requires the computation of a 3D matrix multiplication; this accounts for $O(C \cdot L^6)$ operations. The tensor products are followed by linear self-interactions; they require the computation of $O(L)$ matrices of dimension $O(C\cdot L^2) \times O(C)$ which accounts for $O(C^2\cdot L^3)$ operations. This becomes prohibitively expensive even with GPUs when we scale $L$.

Our $SO(2)$ generalized convolutions can be efficiently parallelized in GPU by computing the matrix multiplications between $\tilde{\mathbf{h}}_{m_o} \in \mathbb{R}^{(L-m_o + 1) \times (L-m_o + 1)}$ and $(\mathbf{D}_{st} \cdot \mathbf{x}_s)_{m_o} \in \mathbb{R}^{L-m_o + 1}$ for $-L \le m_o \le L$. 
That is, we need $O(L)$ 2D matrix multiplications to compute the full tensor product. This accounts for $O(C^2 \cdot L^3)$ operations. Moreover we need an additional $O(C \cdot L^3)$ operations to rotate the various irreps to align the edge's direction to the y-axis.
A similar analysis shows that the GPU memory allocated at training time scales with $O(C^2 \cdot L^3)$ for the $SO(2)$ formulation, whereas the naive implementation is $O(C \cdot L^6 + C^2 \cdot L^3)$.

In Figure \ref{fig:e3nn_comp}, we empirically validate these results by comparing the training time per epoch and the GPU memory allocated at training time over $L$. We fix the number of hidden channels to $64$, remove any non-linearity in the network and train a pair of mathematically equivalent GNNs for $1 \le L \le 5$.

\section{Analysis on the quasi-equivariance of the spherical activation function}
\label{app:quasi}
\begin{figure}[h]
\begin{center}
    \includegraphics[width=0.6\textwidth]{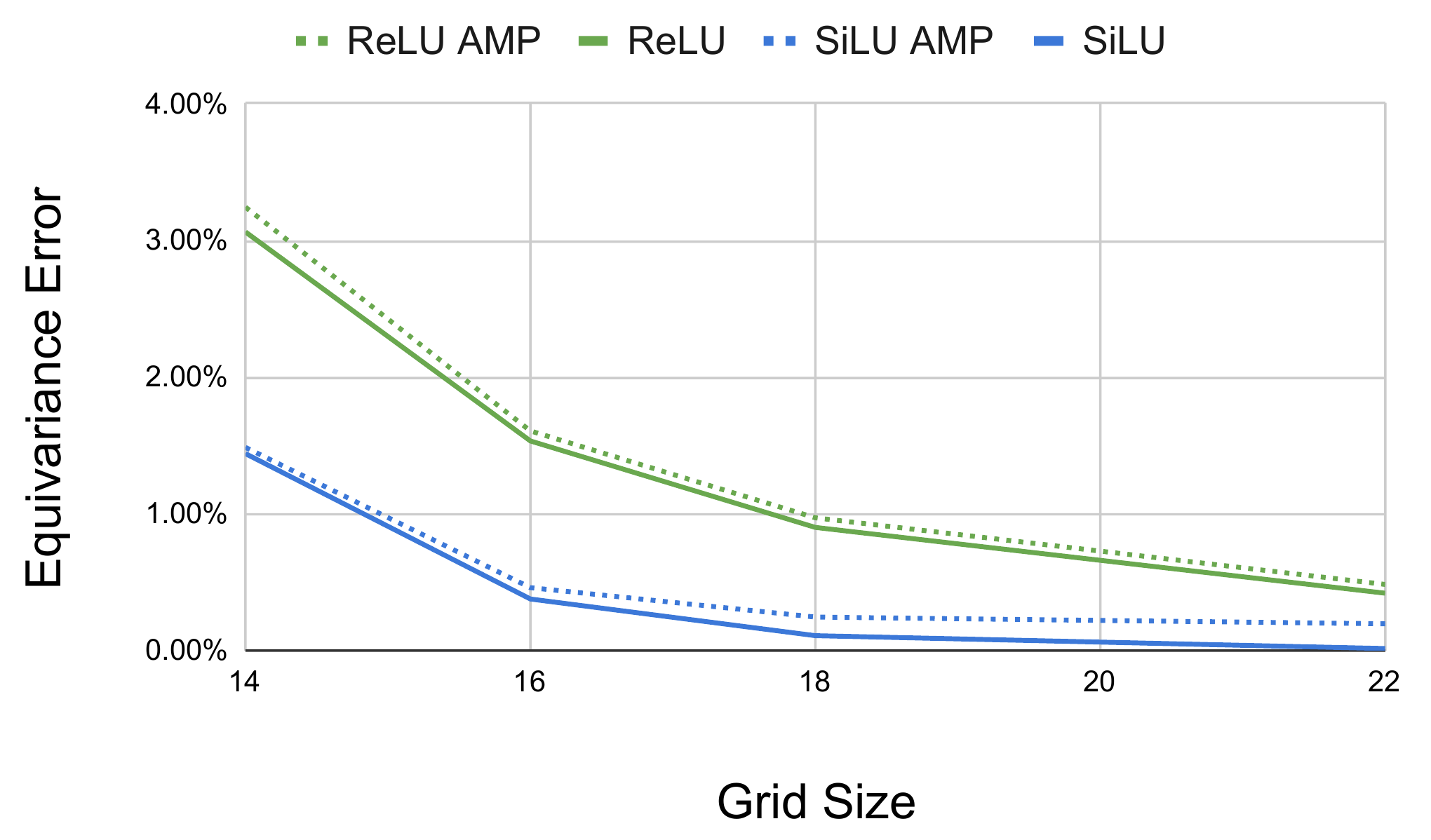}
\vspace{-6pt}
\caption{Plot of the percentage error between the calculation of two rotated messages as the grid size of the point-wise non-linear activation function (SiLU or ReLU) is increased. Plots are shown with and without the use of AMP. The model has 12 layers and $L=6$. A grid size of 14 with SiLU and AMP was used in the paper which results in a $1.5\%$ relative error. However, larger grid sizes may be used to reduce the error to a value close to zero if AMP is not used. Since AMP uses lower precision, the best percentage error is limited to $0.2\%$ achieved at a grid size of 18, i.e., the error due to AMP overwhelms any error due to equivariance errors. If a grid size of 18 is used instead of 14, no improvement in energy or force MAE was observed.}
\label{fig:MsgEquiv}
\end{center}
\end{figure}
\model~introduces non-linearities through a point-wise non-linearity described in Equations (\ref{eqn:pointwise}) and (\ref{eqn:pointwise-aggr}). 
\citep{cohen2016group} proves that point-wise non-linearities are equivariant. However, the numerical approximation of the integration with a discrete number of samples results in a small loss of equivariance per-layer. Equation (\ref{eqn:pointwise}) is approximated by sampling points on a uniformly spaced grid along the directions $(\theta, \phi)$ with dimension $(2\cdot M + 1) \times (2 \cdot L + 1)$. Equation (\ref{eqn:pointwise-aggr}) is sampled with resolution $(2\cdot L + 1) \times (2 \cdot L + 1)$. If the activation function was linear, these resolutions would result in perfect equivariance. However, the higher frequencies introduced by a non-linear activation function can result in aliasing.

While not being perfectly equivariant, point-wise non-linearities have been adopted in many quasi-equivariant models, i.e., models with an equivariance error close to zero \citep{cohen2018spherical}. We study the equivariance error of the spherical activation of Equation (\ref{eqn:pointwise}) by varying the longitudinal and latitudinal grid size and the non-linear function employed. We measure the average percentage error between two rotated messages by:
\begin{equation}
    eq. \text{ }err(\mathbf{a}_{st}) = \mathbb{E}\left[\left| \int \mathbf{Y} (\hat{\mathbf{r}}) P(F_{\mathbf{a}_{st}}(\hat{\mathbf{r}})) \, d\hat{\mathbf{r}} -  \mathbf{D}(R^{-1}) \int \mathbf{Y} (\hat{\mathbf{r}}) P(F_{\mathbf{D}(R)\mathbf{a}_{st}}(\hat{\mathbf{r}})) \, d\hat{\mathbf{r}} \right|\right]
\end{equation}
where $P$ is a non-linear function chosen between $ReLU$ and $SiLU$ and the expectation is taken over uniformly sampled $R \in SO(3)$. For the settings used in this paper with a grid size of 14 with SiLU and AMP, only a small equivariance error ($1.5\%$) is observed. This error may be reduced below the level observable with AMP using a grid size of 18. Note that other activation functions that introduce more higher frequencies, such as ReLU, may require higher resolutions if numerically perfect equivariance is desired. 
\section{Sample efficiency of \model}\label{app:sample}
\begin{figure}
  \centering
  \includegraphics[width=0.5\textwidth]{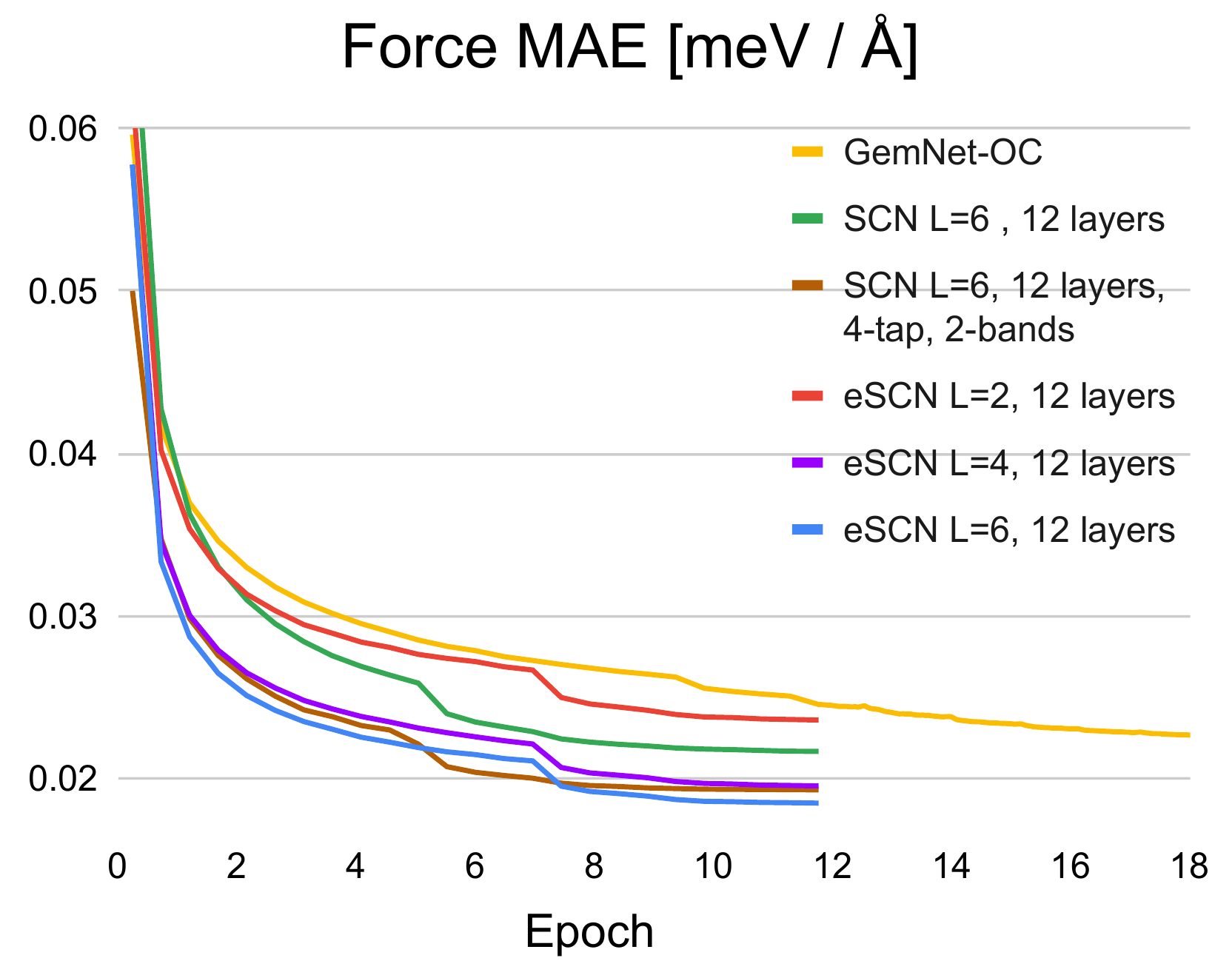}
    \vspace{-6pt}
  \caption{Training curves for several model variants on the OC20 2M dataset. Force Mean Absolute Errors (MAEs) are computed on a 30k subset of the Validation ID set in OC20. Note that \model~has high sample efficiency, similar to SCN.} \label{fig:sample}
\end{figure}

We provide an additional study on the sample efficiency of \model~by empirically comparing it to state-of-the-art GemNet-OC \citep{gasteiger2022gemnet} and SCN \citep{zitnick2022spherical}. In Figure \ref{fig:sample} we plot the Force MAE by training Epochs on the OC-20 2M \citep{OC20} dataset. We conclude that \model~preserves the same sample efficiency of SCN.

\end{document}